\documentclass[letterpaper, 10 pt, conference]{ieeeconf}
\pdfoutput=1

\IEEEoverridecommandlockouts
\makeatletter
\let\NAT@parse\undefined
\makeatother

\usepackage{floatrow} 
\usepackage[utf8]{inputenc}
\usepackage{pgfplots}
\usepackage{color}
\usepackage{mathrsfs,amsmath,amssymb,amsfonts}
\usepackage{mathtools,bbm,bm}
\usepackage{graphicx}
\graphicspath{{./fig/}}
\usepackage[
hidelinks,
colorlinks,
linkcolor=black,
citecolor=black,
filecolor=red,
urlcolor=blue
]{hyperref}
\usepackage{moreverb}
\usepackage{setspace}
\usepackage{url}
\usepackage{tikz}
\usepackage{subcaption}
\usepackage[font=footnotesize,labelfont=bf]{caption}
\usepackage[numbers,sort&compress]{natbib}
\usepackage{algorithm}
\usepackage[noend]{algpseudocode}
\usepackage{xcolor}
\usepackage[bottom,multiple]{footmisc}
\usepackage{physics} 
\usepackage{xparse} 
\usepackage{balance}

\pgfplotsset{plot coordinates/math parser=false}

\pgfplotsset{compat=1.13} 

\DeclareGraphicsExtensions{.pdf,.png,.jpg}

\newlength\figureheight
\newlength\subfiguresheight
\newlength\figurewidth
\newlength\phantomheight
\newlength\compositewidth

\newtheorem{theorem}{Theorem}
\newtheorem{definition}{Definition}

\newtheorem{lemma}[theorem]{Lemma}
\newtheorem{remark}{Remark}

\definecolor{cmu_red}{rgb}{0.6, 0.0, 0.0}

\newif\ifshowdevelopment

\showdevelopmentfalse

\newcommand{\todo}[1]{%
  \ifshowdevelopment%
    \textcolor{blue}{\emph{\bf TODO:\ #1}}%
  \else\fi%
}

\newif\ifextended
\extendedtrue

\DeclareMathOperator*{\argmax}{arg\,max}

\newcommand{\spacefont}[1]{{\mathbb{#1}}}

\renewcommand{\real}{{\spacefont{R}}}
\newcommand{\integer}{{\spacefont{Z}}}

\renewcommand{\vec}[1]{\mathbf{#1}}

\newcommand{\probabilityfont}[1]{{\mathbb{#1}}}
\newcommand{\E}{{\probabilityfont{E}}}

\newcommand{\MI}{{\probabilityfont{I}}}

\newcommand{\setsystemfont}[1]{{\mathscr{#1}}}
\newcommand{\ground}{\Omega}
\newcommand{\independence}{\setsystemfont{I}}
\newcommand{\block}{\setsystemfont{U}}

\newcommand{\setfun}{g}
\newcommand{\setfunsim}{\setfun^\mathrm{sim}} 

\newcommand{\setfuns}{\mathscr{G}} 

\newcommand{\graphfont}[1]{{\mathcal{#1}}}
\newcommand{\graph}{{\graphfont{G}}}
\newcommand{\edges}{{\graphfont{E}}}
\newcommand{\weights}{{\graphfont{W}}}
\newcommand{\approxweights}{{\graphfont{\widehat W}}}

\newcommand{\opt}{{\star}}

\newcommand{\neighbor}{\mathcal{N}}
\newcommand{\targets}{\mathcal{T}}

\newcommand{\numtargets}{{n_\mathrm{t}}}
\newcommand{\numrobots}{{n_\mathrm{r}}}
\newcommand{\numrounds}{{n_\mathrm{d}}}
\newcommand{\robots}{{\mathcal{R}}}

\newcommand{\method}[1]{{\textsc{#1}}}
\newcommand{\inneighbor}{\neighbor^\mathrm{in}}
\newcommand{\outneighbor}{\neighbor^\mathrm{out}}
\newcommand{\ignore}{{\widehat{\mathcal{N}}}^\mathrm{in}}
\newcommand{\data}{\theta}
\newcommand{\approxsetfun}{\widetilde\setfun}

\newcommand{\cost}{\gamma}
\newcommand{\costgeneral}{\gamma^\mathrm{gen}}
\newcommand{\objectivecost}{\cost^\mathrm{obj}}
\newcommand{\plannercost}{\cost^\mathrm{plan}}
\newcommand{\distcost}{\cost^\mathrm{dist}}

\newcommand{\robotstates}{\vec{X}^\mathrm{r}}
\newcommand{\targetstates}{\vec{X}^\mathrm{t}}
\newcommand{\robotposition}{\vec{p}^\mathrm{r}}
\newcommand{\targetposition}{\vec{p}^\mathrm{t}}
\newcommand{\observations}{\vec{Y}}

\newcommand{\robotstate}{\vec{x}^\mathrm{r}}
\newcommand{\targetstate}{\vec{x}^\mathrm{t}}

\newcommand{\robotspace}{\real^{d^\mathrm{r}}}
\newcommand{\targetspace}{\real^{d^\mathrm{t}}}

\newcommand{\robotdynamics}{f^\mathrm{r}}
\newcommand{\targetdynamics}{f^\mathrm{t}}

\newcommand{\control}{u}
\newcommand{\controlspace}{\mathcal{U}}

\newcommand{\observation}{\vec{y}}
\newcommand{\observationfunction}{h}

\newcommand{\targetnoise}{\epsilon^\mathrm{t}}
\newcommand{\targetnoisespace}{\mathcal{E}^\mathrm{t}}
\newcommand{\observationnoise}{\epsilon^\mathrm{y}}

\newcommand{\capacity}{C}

\newcommand{\robotmap}{\Phi^\mathrm{r}}
\newcommand{\targetmap}{\Phi^\mathrm{t}}
\newcommand{\locationspace}{\real^3}

\newcommand{\robotmaxtravel}{d^\mathrm{r}}
\newcommand{\targetmaxtravel}{d^\mathrm{t}}

\definecolor{endcolor}{RGB}{196,18,48}
\definecolor{differencecolor}{RGB}{0,45,114}

\newcommand{\optionalsubscript}[1][]{%
  \ifthenelse{\equal{#1}{}}{%
  }{%
    $_{#1}$%
  }%
}

\NewDocumentCommand{\distalgnamed}{O{} O{} m}{%
  \ifthenelse{\equal{#2}{}}{%
    {\text{\textnormal{#3}\optionalsubscript[#1]{}}}%
  }{%
    {\text{#2{#3}\optionalsubscript[#1]{}}}%
  }%
}

\NewDocumentCommand{\rsp}{O{} O{}}
{%
  {\distalgnamed[#1][#2]{RSP}}%
}
\NewDocumentCommand{\rrsp}{O{} O{}}
{%
  {\distalgnamed[#1][#2]{R-lRSP}}%
}

\begin{document}

\title{
  Scalable Distributed Planning for Multi-Robot, Multi-Target Tracking
}

\author{%
Micah Corah
and
Nathan Michael
\thanks{
  Micah Corah is affiliated with the NASA Jet Propulsion Laboratory (JPL),
  Pasadena, CA, USA.
  Nathan Michael is affiliated with the Robotics Institute, Carnegie Mellon
  University (CMU), Pittsburgh, PA, USA.
  This work was completed while Micah Corah was a student at CMU.
}
\thanks{
  \tt{mcorah@jpl.nasa.gov, nmichael@cmu.edu}%
}
}

\maketitle

\begin{abstract}
  In multi-robot multi-target tracking, robots coordinate to monitor
  groups of targets moving about an environment.
  We approach planning for such scenarios by formulating a
  receding-horizon, multi-robot sensing problem with a mutual information
  objective.
  Such problems are NP-Hard in general.
  Yet, our objective is submodular which enables
  certain greedy planners to guarantee constant-factor suboptimality.
  However,
  these greedy planners require robots to plan their actions
  in sequence, one robot at a time, so
  planning time is at least proportional to the number of robots.
  Solving these problems becomes intractable for large teams, even for
  distributed implementations.
  Our prior work proposed a distributed planner (\rsp{}) which reduces this
  number of sequential steps to a constant, even for large numbers of robots, by
  allowing robots to plan in parallel while ignoring some of each others'
  decisions.
  Although that analysis is not applicable to target tracking,
  we prove a similar guarantee, that \rsp{} planning approaches performance
  guarantees for fully sequential planners, by employing a
  novel bound which takes advantage of the independence of target motions
  to quantify effective redundancy between robots' observations and actions.
  Further, we present analysis that explicitly accounts for features of
  practical implementations including approximations to the objective and
  anytime planning.
  Simulation results---available via open source release---for target tracking
  with ranging sensors demonstrate that our planners consistently approach the
  performance of sequential planning (in terms of position uncertainty) given
  only 2--8 planning steps and for as many as 96 robots with a 24x
  reduction in the number of sequential steps in planning.
  Thus, this work makes planning for multi-robot target tracking
  tractable at
  much larger scales than before, for practical planners and general tracking
  problems.
\end{abstract}

\section{Introduction}

In target tracking problems, robots seek to observe a number of discrete targets
whose states may evolve in time,
such as for surveillance, monitoring wildlife~\citep{cliff2015rss}, and
intercepting rogue UAVs~\citep{shah2019ras}.
The robots may track the positions of the targets as well as other features of
their states, such as to track an animal's actions or the pose of an elite
athlete.
When a large number of such targets are spread over more space than a
single robot can cover, deploying more robots can improve their capacity to
track the targets but at the expense of more complex planning problems.

Still, even simple target tracking problems can be difficult to model and
solve.
Noisy range observations can
produce multi-modal posterior distributions over target positions which do
not have closed-form solutions;
and planning and tracking systems often approximate both filter posteriors and
sensing utility~\citep{charrow2014auro,charrow2014ijrr}.
Likewise, realistic environments induce complexity by constraining target motion
such as for
search on a road network~\citep{piacentini2019jair} or in an indoor office
space~\citep{hollinger2009ijrr}.
Additionally, deep learning methods for visual object tracking enable
systems to track wide varieties of objects, and these objects may have similarly
varied states and dynamics~\citep{wang2019cvpr}.
This motivates development of objectives and planners that capture the
nuances of these problems.
To address this, we provide analysis for general tracking problems
where target states are independent of each other and the robots tracking
them\footnote{%
  Although problems may be adversarial in other parameters, this excludes
  pursuit-evasion problems~\citep{chung2011auro,shah2019ras}
  where targets can observe pursuers.
}
with a flexible objective (mutual information), and
we allow for choice of approximate representations and planners.

Systems for target tracking in multi-robot settings often rely
on greedy algorithms~\citep{tokekar2014iros,zhou2019tro} which
apply to a wide variety of relevant submodular
objectives, including the mutual information
objectives we study~\citep{krause2005uai,schrijver2003}.
Additionally, these greedy algorithms can augment general single-robot planners
to provide efficient planning and constant-factor performance guarantees in
multi-robot domains~\citep{singh2009}
despite common formulations of these problems being
NP-Hard~\citep{krause2005uai,feige1998}.

However, a limitation of greedy algorithms for multi-robot planning---especially
in distributed settings---is that robots must make decisions
sequentially~\citep{atanasov2015icra} so that planning time grows with the
number of robots, and this growth in planning time can prohibit large teams of
robots from reacting promptly to target motions.
Advances in distributed algorithms relevant to target tracking
begin to address this issue but are currently limited to coverage-like
objectives~\citep{corah2018cdc,sung2019auro}.
Here, we extend our analysis for \rsp{}
planning~\citep{corah2018cdc,corah2020phd} to include more general mutual
information objectives.
Analysis for sequential planners also typically assumes individual robots obtain
solutions that are either exact~\citep{fisher1978,atanasov2015icra} or within a
constant factor of optimal~\citep{singh2009,jorgensen2017iros,corah2019auro}.
Although such analysis is appropriate for single-robot planners with guarantees
on solution
quality~\citep{chekuri2005,singh2009,jorgensen2017iros}, assuming
constant-factor suboptimality is less suited for
anytime or sampling-based
planners~\citep{hollinger2014ijrr,lauri2015ras,atanasov2015icra}
like those we apply in this paper.

\subsection{Contributions}
This paper extends methods for distributed planning to target
tracking problems, presents analysis that accounts
for common approximations, and applies these methods to multi-robot planning in
a simulated target tracking scenario.\footnote{
  This work and~\citep{corah2018cdc} appear as chapters
  in the thesis~\citep{corah2020phd}.
}

\subsubsection{Distributed planning for target tracking}
Although sequential planners generally require computation time that is at least
proportional to the number of robots, recent works on distributed optimization
introduced methods that can reduce the number of sequential
steps~\citep{gharesifard2017,grimsman2018tcns}.
Our prior works built on these to develop planners based on
Randomized Sequential Partitions\footnote{%
  Although \citep{corah2018cdc} introduces the method,
  \citep{corah2020phd} introduces the term \rsp{}.
}
(\rsp{})
that run in constant numbers of steps, independent of the
number of robots~\citep{corah2018cdc,corah2020phd}.
While the prior analysis is only relevant to coverage objectives,%
\footnote{%
  \citet[Fig.~1]{corah2018cdc} also describe a different mutual information
  objective which violates requirements for ``coverage-like'' objectives.
}
this paper demonstrates that \rsp{} planning is applicable to
target tracking with mutual information objectives by providing guarantees on
solution quality in terms of a bound on the effective pairwise redundancy
between robots' actions.
We obtain this bound after decomposing the objective as a sum of submodular
functions over each target.
This analysis demonstrates that distributed planners running in constant time
can guarantee performance approaching that of more inefficient sequential
planners (within half of optimal)~\citep{fisher1978}.

\subsubsection{Analysis of approximate, anytime planning}
The analysis of \rsp{} also introduces a novel approach to account
for common sources of suboptimality in receding-horizon planning
(due to \emph{approximation of the objective} and
\emph{suboptimal single-robot planning}).
This affirms that methods for submodular maximization are applicable to
target tracking in the presence of approximate objective values and anytime
planners that may sometimes produce poor results or fail.

\subsubsection{Application to a multi-robot multi-target tracking problem}
Finally, we apply the analysis to develop a planner\footnote{%
  Source code for the target tracking simulations is available at:
  \url{https://github.com/mcorah/MultiAgentSensing}
}
for multi-robot
multi-target tracking with a mutual information objective and
demonstrate that a distributed \rsp{} planner running in constant time can
guarantee suboptimality approaching that of fully sequential planning.
Then, additional simulation results confirm that \rsp{} maintains
consistent solution quality for up to 96 robots.
This produces a $24\!\times$ reduction in the number of sequential planning
steps and an at least equivalent reduction in planning duration.

\section{Background}

Let us begin by presenting the mathematical background on our approach to
multi-robot planning.

\subsection{Set functions and their properties}
\label{sec:set_functions}
Functions of sets $\setfun : 2^\ground \rightarrow \real$ can quantify the
utility of sets of control actions, each a subset of a finite collection
of possible actions $\ground$, and
we seek to maximize set functions that satisfy the following conditions.
First, $\setfun$ is normalized if $\setfun(\emptyset)\!=\!0$.
Additionally, objectives in sensing tasks~\citep{%
  krause2008jmlr,hollinger2009ijrr,singh2009,atanasov2015icra%
}
often have useful monotonicity properties when written as set functions.
A function is monotonically increasing if
\begin{align}
  \setfun(A) \geq \setfun(B)
  \label{eq:monotonicity}
\end{align}
where $B \subseteq A \subseteq \ground$.
Next, functions with monotonically decreasing
marginal gains are \emph{submodular}, satisfying the following
inequality
\begin{align}
  \setfun(A \cup C) - \setfun(A) \leq \setfun(B \cup C) - \setfun(B)
  \label{eq:submodularity}
\end{align}
where $C \in \ground \setminus A$.
The differences in \eqref{eq:submodularity} express
discrete derivatives of $\setfun$.
Drawing on notation for mutual information~\citep{cover2012}, the
$n^\mathrm{th}$ derivative of $\setfun$ at $X \subseteq \ground$ with respect to
disjoint sets $Y_1,\ldots Y_n \subseteq \ground$
can be defined recursively as
\begin{align}
  \setfun(Y_1;\ldots;Y_n | X) &=
  \setfun(Y_1;\ldots;Y_{n-1}|X \cup Y_n) \nonumber\\
  &\hspace{2em} - \setfun(Y_1;\ldots;Y_{n-1}|X \cap Y_n)
  \label{eq:derivative}
\end{align}
where $\setfun(X)=\setfun(\cdot|X)$ is the $0^\mathrm{th}$ derivative.
So, \eqref{eq:submodularity} can be written as
$\setfun(C|A)\leq\setfun(C|B)$, and so
both monotonicity~\eqref{eq:monotonicity} and
submodularity~\eqref{eq:submodularity} form monotonicity conditions on
derivatives of $\setfun$~\citep{foldes2005,corah2020phd}.
Further, second derivatives are written as\footnote{
  We ignore intersections in \eqref{eq:derivative} as variables are disjoint.
}
$\setfun(A;B|C) = \setfun(A|B\cup C) - \setfun(A|C)$ which expresses
\emph{effective redundancy} between $A$ and $B$.

This text also abuses notation for sets and set functions:
writing arguments to set functions like multivariate functions
so
$\setfun(A,B)\!=\!\setfun(A \cup B)$;
implicitly wrapping elements $x\!\in\!\ground$ in sets
$\setfun(x)\!=\!\setfun(\{x\})$;
writing integer ranges with the notation
$i\!:\!j = \{k \mid i \leq k \leq j,\ k\in\integer\}$; and
indexing into sets with sets of integers in subscripts as in $A_{1:5}$ where
intent is clear.

\subsection{Partition matroids for multi-robot systems}
\label{sec:partition_matroids}

Each robot $i\in\robots$ in a team
$\robots = \{1,\ldots,\numrobots\}$ has access to a unique set of local control
actions $\block_i$ (i.e. the set of feasible receding-horizon trajectories).
These sets of control actions are disjoint and together form
(and partition) the set of all available actions
$\ground = \bigcup_{i\in\robots} \block_i$.
Each robot may choose any one action from its local set, and the set of all such
complete and incomplete assignments forms a simple partition matroid
$\independence=
\{X \subseteq \ground \mid 1 \geq |X \cap \block_i|, \forall i\in
\robots\}$~\citep[Sec.~39.4]{schrijver2003}.

\section{Target tracking problem}

\begin{figure}
  \floatbox[{\capbeside\thisfloatsetup{capbesideposition={right,top},capbesidewidth=0.37\linewidth}}]{figure}[0.575\textwidth]
  {\caption{%
      A team of aerial robots $\robots$ (black) plan over a receding horizon to
      track a number of targets $\targets$ (red).
      The
      robots select sensing actions to minimize uncertainty in the
      target states which evolve independently of each other and the robots.
  }%
  \label{fig:target_tracking}
  }
  {\includegraphics[width=\linewidth]{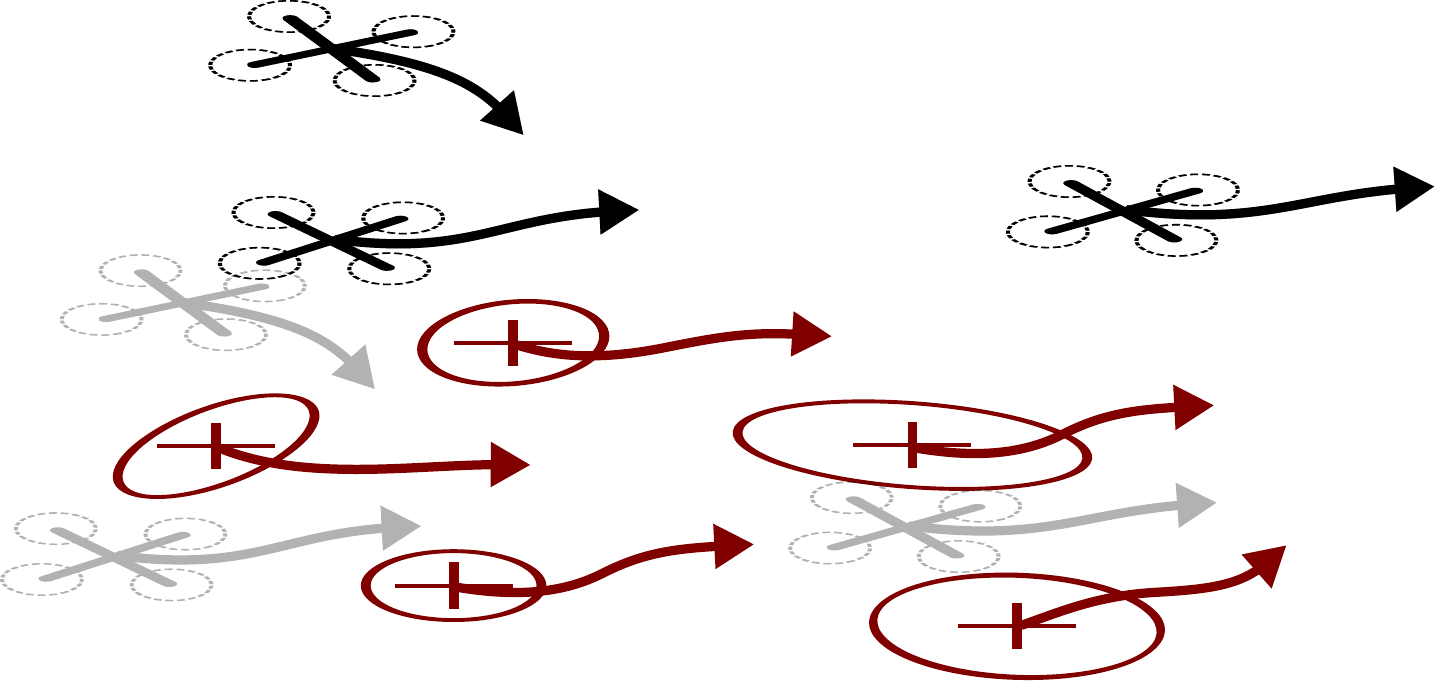}}
\end{figure}

Consider a set of moving targets $\targets=\{1,\ldots,\numtargets\}$
and robots tracking those targets $\robots=\{1,\ldots,\numrobots\}$, seeking to
minimize uncertainty (i.e. entropy~\citep{cover2012}), as illustrated in
Fig.~\ref{fig:target_tracking}.
Let $\robotstate_{i,t}\in\robotspace$ and
$\targetstate_{j,t}\in\targetspace$ be the respective states of robot
$i\in\robots$ and target $j\in\targets$ at time $t\in\{0,\ldots,T\}$.
The states of each evolve in discrete time, with known dynamics
\begin{align}
  \robotstate_{i,t+1} &= \robotdynamics(\robotstate_{i,t},\control_{i,t}),
    &
  \targetstate_{j,t+1} &=
  \targetdynamics(\targetstate_{j,t},\targetnoise_{j,t}),
  \label{eq:dynamics}
\end{align}
where $\control_{i,t}\in\controlspace$ is a control input from a finite set of
inputs $\controlspace$
and $\targetnoise_{i,t}$ is
the targets' process noise.
The robots then receive noisy observations $\observation_{i,j,t}$ of the target
states via
\begin{align}
  \observation_{i,j,t} &=
  \observationfunction(
  \robotstate_{i,t},\targetstate_{j,t},\observationnoise_{i,j,t})
  \label{eq:observations}
\end{align}
where $\observationnoise_{i,j,t}$ is the observation noise.
We refer to states and observations collectively with boldface capitals
as $\robotstates_t$, $\targetstates_t$, and $\observations_t$, each at
time $t$.

\subsection{Receding-horizon optimization problem}
\label{sec:optimization_problem}

Every so often, the robots plan to jointly maximize information gain over a
receding horizon, starting at time $t$ and with duration $l$, in what we refer to
as a planning \emph{epoch}.
Specifically, robots maximize a submodular, monotonic, normalized objective
$\setfun$ subject to a partition matroid constraint so that the optimal set of
control actions is
\begin{align}
  X^\opt \in \argmax_{X\in\independence} \setfun(X).
  \label{eq:optimization_problem}
\end{align}
The partition matroid $\independence$ (Sec.~\ref{sec:partition_matroids})
represents assignment of sequences of control actions to robots with local sets
\begin{align}
  \block_i &= \{(i, \control_{1:l})\mid \control_{1:l}\in \controlspace^l\}
  \quad \forall i\in\robots.
\end{align}
The objective $\setfun$ is the mutual information between observations and
target states given the choice of control actions.
Now, interpret future states
$\targetstates_{t+1:t+l}$
and observations
$\observations_{t+1:t+l}$
(taking care not to confuse states $\vec{X}$ with sets $X$)
as random variables
induced by the process~\eqref{eq:dynamics} and
observation~\eqref{eq:observations} noise terms.
Then, writing future observations\footnote{%
  $\observations_{t+1:t+l}(X)$ is a random variable expressing
  the noisy observations \eqref{eq:observations} that robot $i$ receives after
  executing $\control_{1:l}$ for each $(i, \control_{1:l}) \in X$.
  For the purpose of analysis, $X\subseteq\ground$ may include multiple
  assignments to one robot which will be associated
  with unique observation noise terms $\observationnoise$.
}
as
$\observations_{t+1:t+l}(X)$ for $X\subseteq\ground$,
the mutual information is
\begin{align}
  \setfun(X) = \MI(\targetstates_{t+1:t+l}; \observations_{t+1:t+l}(X)
  |\observations_{0:t},\robotstates_{0:t})
  \label{eq:objective}
\end{align}
where $\MI(X;Y|Z)$ is the Shannon mutual information between $X$ and $Y$
conditional on $Z$ and quantifies the reduction in uncertainty (entropy) of one
random variable from observing another.
We refer interested readers to~\citet{cover2012} for more detail and thorough
definitions.
Crucially, mutual information objectives are normalized, monotonic, and
submodular (defined in Sec.~\ref{sec:set_functions}) when observations
are conditionally independent of target states~\citep{krause2005uai}
as they are here.
However, mutual information does not satisfy the higher-order monotonicity
condition which our prior work
employs~\citep{corah2018cdc}.
Instead, this work takes advantage of properties of a factored form of the
objective.
Equation \eqref{eq:objective} can be factored as a sum over the targets
as
\begin{align}
  \setfun(X) =
  \sum_{j\in\targets}
  \MI(\targetstates_{j,t+1:t+l};
  \observations_{j,t+1:t+l}(X) |\observations_{j,0:t},\robotstates_{0:t})
  \label{eq:objective_sum}
\end{align}
because the
target states and observations of the same are jointly independent and because
the robot dynamics are deterministic.
This follows because the mutual information is a difference of
entropies~\citep[Eq. 2.45]{cover2012} which each decompose as sums over
targets~\citep[Theorem 2.6.6]{cover2012}.

\todo{The last two parts of this section are a bit out of place}

\subsection{Channel capacities and spatial locality}
\label{sec:spatial_locality}

Spatial locality in target tracking problems arises from the factored form
of the objective and how robots' capacities to sense targets decrease
with distance.
Variations in robots' $i\in\robots$ abilities to sense different targets
$j\in\targets$ produce this spatial locality which take the form of channel
capacities $\capacity_{i,j}$ from information
theory~\citep[Chapter~7]{cover2012} so that
\begin{align}
  C_{i,j} =
  \max_{x \in \block_i}
  \MI(\targetstates_{j,t+1:t+l};
  \observations_{j,t+1:t+l}(x) |\observations_{j,0:t},\robotstates_{0:t}).
  \label{eq:target_capacities}
\end{align}
This channel capacity is an \emph{upper bound on the amount of information a
robot may obtain about a given target.}
These channel capacities themselves form informative planning problems
which we solve for the simulation results.

\subsection{Computational model}

A common feature of distributed planning problems is limited
access to information.
We assume for our computational model that each robot $i \in
\robots$ is able to approximate the objective for its own set of actions
$\block_i$.
That is, each robot has access to an approximation of marginal gains
$\approxsetfun_i(x_i | A)$ which is valid for only its own actions $x_i \in
\block_i$ and given any prior selections $A \subseteq \ground$.
This expresses both how robots evaluate mutual information
approximately and how
limited access to sensor data for distant targets could prohibit accurate
evaluation of marginal gains for distant robots.

Robots likewise have limited access to the set of all control actions $\ground$:
robots do not have access to others' actions except those they obtain by
communicating each others' decisions, and
robots also only obtain elements of their local sets $\block_i\subseteq\ground$
implicitly via local planners.

\section{Approach to distributed planning}

The proposed distributed planning framework seeks to approximate
sequential solvers that are known to be near-optimal~\citep{fisher1978} but
become impractical at large scales.
This motivates our distributed approach which is similar but utilizes
modifications and approximations to achieve planning in constant time and to
satisfy constraints on information access for online, receding horizon planning.
Later, Sec.~\ref{sec:cost_model} will describe the costs of these
approximations, and Sec.~\ref{sec:analysis} will integrate these costs into a
suboptimality guarantee that relates our distributed planner
to sequential planning.


\subsection{Greedy, sequential planning (an idealized planner)}
\label{sec:local_greedy}

The local greedy algorithm, originated by \citet{fisher1978} and
applied to robotics by \citet{singh2009} enables robots to obtain
near-optimal solutions to sensing problems by planning in sequence
conditional on prior decisions.
Robots produce solutions
$X^\mathrm{g}=\{x^\mathrm{g}_1,\ldots,x^\mathrm{g}_\numrobots\}$
by greedy planning
\begin{align}
  x^\mathrm{g}_i \in \argmax_{x\in\block_i} \setfun(x|X_{1:i-1}^\mathrm{g}).
  \label{eq:sequential_greedy}
\end{align}
This local greedy algorithm is specific to partition matroids
(Sec.~\ref{sec:partition_matroids})
and differs subtly from greedy algorithms
for general matroids~\citep{fisher1978,choi2009tro,williams2017icra}
which maximize over all possible actions ($\ground$)
at each greedy step rather than robots' local control actions ($\block$) as in
\eqref{eq:sequential_greedy}.
Still, both greedy algorithms satisfy a
constant-factor bound
$\setfun(X^\mathrm{g}) \geq \frac{1}{2}\setfun(X^\opt)$.
While the local algorithm completes only one complete pass over
$\ground$ rather than one per robot,
both require robots to make decisions in sequence:
\emph{the duration of planning for \eqref{eq:sequential_greedy} is at least
proportional to the size of the team.}

\subsection{Distributed planning algorithm}

\begin{algorithm}[b]
  \caption{Distributed algorithm for receding-horizon target tracking from the
  perspective of robot $i\in\robots$ for execution at time
  $t$.}\label{alg:distributed}
  \begin{minipage}{\linewidth}
    \begin{algorithmic}[1]
      \State $\inneighbor_i \gets \text{\emph{in}-neighbors of robot } i$
      \State $\outneighbor_i \gets \text{\emph{out}-neighbors of robot } i$
      \State $\data_i \gets \text{sensor data (or summary) accessible to
      robot } i$
      \vskip 1.5ex
      \State \method{Receive}: $X^\mathrm{d}_{\inneighbor_i}$
      from $\inneighbor_i$
      \label{line:receive}
      \State
      $\approxsetfun_i \gets \text{approximation of }
      \setfun \text{ given } \data_i$
      \State $x^\mathrm{d}_i \gets
      \method{PlanAnytime}(\approxsetfun_i,
                           \robotstate_{i,t},
                           X^\mathrm{d}_{\inneighbor_i})$
      \label{line:anytime}
      \State \method{Send}: $x^\mathrm{d}_i$ to $\outneighbor_i$
      \label{line:send}
      \State \method{Execute}: $x^\mathrm{d}_i$ starting at time $t$ and until
      the beginning of the result of the next planning round
      \label{line:execute}
    \end{algorithmic}
  \end{minipage}
\end{algorithm}

\begin{figure}
  \newcommand{\quadblack}{\includegraphics[width=0.7cm]{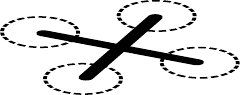}}
  \centering
  \begin{subfigure}[t]{0.60\linewidth}
    \centering
    \begin{tikzpicture}[xscale=1.4]
      \tikzset{vertex/.style = {shape=circle,draw,minimum size=1.5em}}
      \tikzset{edge/.style = {->,> = latex, ultra thick}}

      \node (1) at (0,0) {\quadblack};
      \node (2) at (1,0) {\quadblack};
      \node (3) at (2,0) {\quadblack};
      \node (4) at (3,0) {\quadblack};

      \draw[edge] (1) to                               (2);
      \draw[edge] (1) to[bend left=50, looseness=1.0] (3);
      \draw[edge] (1) to[bend  left=50, looseness=1.0] (4);

      \draw[edge] (2) to                               (3);
      \draw[edge] (2) to[bend left=50, looseness=1.0] (4);

      \draw[edge] (3) to                               (4);
    \end{tikzpicture}
    \caption{Sequential}\label{subfig:sequential}
  \end{subfigure}%
  \begin{subfigure}[t]{0.40\linewidth}
    \centering
    \begin{tikzpicture}[xscale=1.4]
      \tikzset{vertex/.style = {shape=circle,draw,minimum size=1.5em}}
      \tikzset{edge/.style = {->,> = latex, ultra thick}}

      \node (1a) at (0,0) {\quadblack};
      \node (1b) at (1,0) {\quadblack};

      \draw[edge] (1a) to                (1b);

      \node (2a) at (0,1.3) {\quadblack};
      \node (2b) at (1,1.3) {\quadblack};

      \draw[edge] (2a) to                (2b);

      \draw[edge] (1a) to (2b);
      \draw[edge] (2a) to (1b);

      \draw[edge, draw=cmu_red] (1a) to (2a);
      \draw[edge, draw=cmu_red] (1b) to (2b);
    \end{tikzpicture}
    \caption{Parallel}\label{subfig:parallel}
  \end{subfigure}
  \caption{
    Graph models for two distributed planners where edges represent access
    to robots' prior decisions.
    (\subref{subfig:sequential}) The sequential greedy algorithm~\eqref{eq:sequential_greedy}
    corresponds to a complete directed acyclic graph
    because robots have access to all prior decisions.
    This graph has one sequential step per robot
    (time increases toward the right).
    (\subref{subfig:parallel}) Deleting edges (red) removes temporal constraints
    due to requiring access to prior decisions.
    Here, deleting edges enables pairs of robots to plan in parallel, and the
    planner subsequently requires two sequential steps instead of four.
  }\label{fig:graphs}
\end{figure}
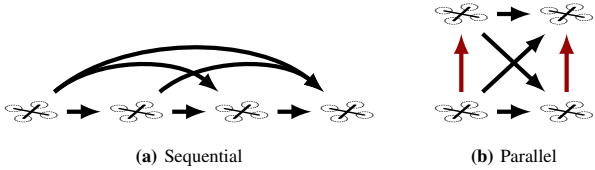

Algorithm~\ref{alg:distributed} provides pseudo-code for a distributed
planner which can run in a constant number of sequential steps and produces
distributed solutions $X^\mathrm{d}$.
As Fig.~\ref{fig:graphs} illustrates, this planner follows a
directed acyclic graph structure where robots are
nodes and an edge to a robot represents access to another's decision.
Designing the graph appropriately---ignoring some decisions---enables
robots to plan in parallel.

Starting in lines~\ref{line:receive}--\ref{line:anytime},
each robot begins planning upon receiving decisions from its in-neighbors
$\inneighbor_i$.
Robots approximate the objective given available computational resources and
sensor data $\data_i$ via $\approxsetfun_i$.
The available sensor data may include data for all targets or only
those near the robot (we will study both cases).
We assume robots obtain this sensor data via inter-robot communication (not
shown).
Then, in lines~\ref{line:send}--\ref{line:execute},
once the planner exits or runs out of time, the robot commits to an action
which it executes in a receding-horizon fashion and
sends to any out-neighbors $\outneighbor_i$.

As described, planning proceeds asynchronously, and robots have access to both
$\inneighbor_i$ and $\outneighbor_i$.
However, \citep[Chapter~8]{corah2020phd} provides a more practical
time-synchronous implementation where the graph structure is implicit.

\subsubsection{Planning in parallel via \rsp[][\textit]{}}
\label{sec:rsp}

To enable parallel planning with little impact on suboptimality,
robots construct the directed graph structure via
Randomized Sequential Partitions (\rsp{})~\citep{corah2018cdc}.
When planning via \rsp{} robots assign themselves randomly to one of
$\numrounds$ sequential steps.
Then, robots assigned to the same step plan in parallel with access
(via $\inneighbor_i$)
to some or all decisions from prior steps.

The cost of planning via \rsp{} instead of sequentially
\eqref{eq:sequential_greedy} approaches zero when effective redundancy between
agents is bounded~\citep{corah2018cdc}.
In Sec.~\ref{sec:analysis},
we will obtain a new such bound for target tracking in terms
of a weighted undirected graph
$\graph = (\robots,\edges,\approxweights)$ which connects the robots
with edges $\edges=\{(i,j)|i,j\in\robots, i \neq j\}$
whereas \citep{corah2018cdc} proves that when the optimum is proportional to
the sum of weights:
\begin{align}
  \setfun(X^\star) \propto \sum_{(i,j)\in\edges} \approxweights(i,j)
  \label{eq:sum_of_weights}
\end{align}
distributed planning \emph{with a constant number of steps} guarantees
suboptimality \emph{approaching half of optimal} (with $1/\numrounds$) in
expectation, for any number of robots~\citep[Theorem~3]{corah2018cdc}.

\section{Cost model for approximate planning}
\label{sec:cost_model}

This section describes the costs of planning with directed acyclic
graphs via \rsp{} and of approximations in planning and objective evaluation
given constraints on computation time and information access.
Rather than assume constant-factor suboptimality at each
step~\citep{singh2009}---as would be appropriate if the local planner satisfied
a consistent performance guarantee---we present this flexible cost model to
account for uncertainty arising from planning in real time.
The analysis (Sec.~\ref{sec:analysis}) will integrate these costs for
\emph{distributed planning}, \emph{objective evaluation},
and \emph{anytime (single-robot) planning} into a suboptimality guarantee
that relates the suboptimality of Alg.~\ref{alg:distributed} to the bound for
sequential planning \eqref{eq:sequential_greedy}.

\subsection{General cost of suboptimal decisions for individual robots}

Before describing the specific costs of approximations, let us define a general
cost in terms of the difference between the exact
marginal gain for a greedy decision and the gain for the actual decision
produced by the planner
$x^\mathrm{d}_i$.
Specifically,
given an instance of \eqref{eq:optimization_problem}
with objective $\setfun$ and
a subset of prior decisions $X \subseteq X^\mathrm{d}_{1:i-1}$,
the cost to robot $i\in\robots$ for making a suboptimal
decision $x^\mathrm{d}_i$ is the difference between the utility of that decision
and the true maximum over $\block_i$
\begin{align}
  \costgeneral_i(\setfun, x^\mathrm{d}_i, X) &=
  \max_{x \in \block_i} \setfun(x|X)
  - \setfun(x^\mathrm{d}_i|X).
  \label{eq:single_robot_suboptimality}
\end{align}
This expression will be useful both for defining specific costs and as a tool
for analyzing suboptimality.

\subsection{Cost of distributed planning on directed acyclic graphs}
\label{sec:graph_model}

Planning nominally via the greedy algorithm~\eqref{eq:sequential_greedy},
robot $i\in\robots$ has access to prior decisions by robots
$\{1,\ldots,i-1\}$.
In our approach, (Alg.~\ref{alg:distributed}) robots
only have access to
a subset $\inneighbor_i \subseteq \{1,\ldots,i-1\}$ of these decisions,
which induces a directed acyclic
graph with edges $(j,i)$ for each robot $j\in\inneighbor_i$ whose decision $i$
accesses while planning~\citep{gharesifard2017,grimsman2018tcns}.
In a sense, the robots ignore decisions by
$\ignore_i=\{1,\ldots,i-1\}\setminus\inneighbor_i$, and the cost of doing so is
a second derivative~\eqref{eq:derivative}:
\begin{align}
  \distcost_i &=
  \setfun(x^\mathrm{d}_i|X^\mathrm{d}_{\inneighbor_i})
  - \setfun(x^\mathrm{d}_i|X^\mathrm{d}_{1:i-1})
  =
  - \setfun(x^\mathrm{d}_i;X^\mathrm{d}_{\ignore_i}|X^\mathrm{d}_{\inneighbor_i}).
  \label{eq:distributed_planning_cost}
\end{align}

This cost expresses the effective redundancy between $i$'s decision
and the decisions by $\ignore_i$ which were ignored.
Later, we will upper bound this redundancy in terms of
$\ignore_i$ and eliminate the dependency on $X^\mathrm{d}_{\inneighbor_i}$.

\subsection{Cost of approximate evaluation of the objective}

We assume robots can access relevant data to evaluate the mutual information
objective~\eqref{eq:objective}, either exactly or approximately such as by
ignoring distant targets or via sampling.
Either way, robot $i\in\robots$ has access to a local approximation
$\approxsetfun_i$ of the objective, and the cost of this approximation (treating
stochasticity implicitly) is at most the sum of the maximum over- and
under-approximation of $\setfun$
\begin{align}
  \begin{split}
    \objectivecost_i =&
    \max_{x_1,x_2 \in \block_i}
    \left(
      \approxsetfun_i(x_1|X^\mathrm{d}_{\inneighbor_i}) -
      \setfun(x_1|X^\mathrm{d}_{\inneighbor_i})
    \right.\\
    &\hspace{5em}+
    \left.
      \setfun(x_2|X^\mathrm{d}_{\inneighbor_i}) -
      \approxsetfun_i(x_2|X^\mathrm{d}_{\inneighbor_i})
    \right)
  \end{split}
  \label{eq:objective_cost}
\end{align}
where $x_1$ and $x_2$ are the points where $\approxsetfun_i$
most under- and over-approximates $\setfun$ over the local action set
$\block_i$.

\subsection{Cost of approximate (anytime) single-robot planning}

Selecting sensing actions to maximize information gain for an individual robot
over a finite horizon produces an informative path planning
problem~\citep{chekuri2005,singh2009}.
However, robots have limited amounts of time available and must terminate
planning and transmit results soon enough so that later robots
that depend on those decisions can make their own decisions in time for the
plans go into effect
(at time $t$ in Algorithm~\ref{alg:distributed}).

Although some existing planners provide performance
guarantees~\citep{chekuri2005,singh2009,zhang2016aaai},
designers applying these methods may have to vary replanning rates
or tune problem parameters to satisfy constraints on planning time for operation
in real-time.
On the other hand,  randomized planners~\citep{lauri2015ras,hollinger2014ijrr}
and gradient- and Newton-based trajectory
generation~\citep{charrow2015rss,indelman2014icra}
converge to local or global maxima but typically provide no guarantees
on solution quality before convergence for anytime planning.
Likewise, this paper applies Monte Carlo tree
search~\citep{browne2012,chaslot2010}, a common randomized planner for
single-robot planning (see the results, Sec.~\ref{sec:results}).

We model online planners via their empirical performance at approximating
optimal single-robot solutions conditional on $X^\mathrm{d}_{\inneighbor_i}$
and with the local objective $\approxsetfun_i$
\begin{align}
  \plannercost_i &=
  \costgeneral_i(\approxsetfun_i, x^\mathrm{d}_i, X^\mathrm{d}_{\inneighbor_i}).
  \label{eq:planner_cost}
\end{align}
This captures the inherent uncertainty in anytime planning and enables
us to characterize collective performance in terms of the bulk suboptimality of
single-robot planning.

\todo{This section is still a bit ugly, and planning discussion could go
elsewhere}

\section{Analysis of suboptimality of distributed planning}
\label{sec:analysis}

This section analyzes suboptimality for distributed planning.
Specifically, Alg.~\ref{alg:distributed} achieves a performance bound which
approaches that for sequential planning (Sec~\ref{sec:local_greedy})
with suboptimality arising from the aforementioned costs.

\begin{theorem}[Suboptimality of Alg.~\ref{alg:distributed}]
  \label{theorem:main_bound}
  Considering an instance of \eqref{eq:optimization_problem}, any solution $X^d$
  that Alg.~\ref{alg:distributed} produces satisfies
  \begin{align}
    \setfun(X^\opt)
    &\leq
    2\setfun(X^d)
    +
    \sum_{i\in\robots}
    \left(
      \distcost_i
      + \objectivecost_i
      + \plannercost_i
    \right),
    \label{eq:costs_bound}
  \end{align}
  and the total cost of distributed planning is bounded by
  \begin{align}
    \sum_{i\in\robots} \distcost_i
    \leq \sum_{i\in\robots}\sum_{j\in\ignore_i} \approxweights(i,j)
    \label{eq:weights_bound}
  \end{align}
  where $\approxweights(i,j)$ is an edge weight
  that bounds effective redundancy between pairs of robots for observing the
  same targets (whose expression we provide in following analysis).
  \emph{This sum of weights approaches zero}
  when planning via \rsp{} with increasing numbers of rounds
  $\numrounds$ (Sec.~\ref{sec:rsp}).
\end{theorem}

The proof of Theorem~\ref{theorem:main_bound} is in
\ifextended
Appendix~\ref{sec:proof_of_main_bound}.
\else
Appendix~III of the extended version~\citep{corah2021irosarxiv}.
\fi
We summarize this result
in (Sec.~\ref{sec:summary_of_proof_of_main_bound}) after introducing preliminary
results related to the first
(Sec.~\ref{sec:general_suboptimality_analysis}) and second
(Sec.~\ref{sec:analysis_of_weights})
parts of the theorem.

Regarding the form of Theorem~\ref{alg:distributed}, this bound characterizes
practical implementations of Alg.~\ref{alg:distributed}.
An idealized version of our distributed algorithm would obtain exact
objective values and maxima, and the associated costs $\objectivecost$ and
$\plannercost$ would each be zero.
From this perspective, \eqref{eq:costs_bound} describes how real implementations
may deviate from this ideal and states that suboptimality arises as an
accumulation of individual inefficiencies which can be modeled empirically.

\subsection{General suboptimality in multi-robot planning}
\label{sec:general_suboptimality_analysis}

The following lemma expresses the joint suboptimality of any solution as a
sum of costs of suboptimal decisions.

\begin{lemma}[Suboptimality of general assignments]
  \label{lemma:approximate_suboptimality}
  Given some submodular, monotonic, normalized objective $\setfun$,
  \emph{any} assignment of actions to all robots (a basis)
  $X^\mathrm{d}\in\independence$ on a simple partition matroid satisfies
  \begin{align}
    \setfun(X^\opt) &\leq    2\setfun(X^\mathrm{d})
    + \sum_{i=1}^{\numrobots}
    \costgeneral_i(\setfun, x^\mathrm{d}_i, X^\mathrm{d}_{1:i-1}).
    \label{eq:approximate_suboptimality}
  \end{align}
\end{lemma}

The proof of Lemma~\ref{lemma:approximate_suboptimality} is in
\ifextended
Appendix~\ref{appendix:approximate_suboptimality}.
\else
Appendix~II.
\fi
Observe that if we obtain $X^\mathrm{d}$ via exact sequential maximization, the
summands ($\costgeneral$) of \eqref{eq:approximate_suboptimality} are zero,
and we obtain the original result by \citet{fisher1978}
and likewise for single-robot solvers with constant-factor
suboptimality~\citep[Theorem~1]{singh2009}.


\subsection{Bounding the cost of distributed planning for target tracking
problems}
\label{sec:analysis_of_weights}

This section characterizes the cost of distributed planning
\eqref{eq:weights_bound} in target tracking problems.
We begin by investigating the decomposition of the objective as a sum over
targets to obtain an intermediate bound.
Applying this bound produces the weights $\approxweights$
which relate the cost of ignoring robots during distributed planning
(Sec.~\ref{sec:graph_model})
to the channel capacities \eqref{eq:target_capacities} between the robots and
targets.

\subsubsection{Decomposing objectives as sums}
The objective \eqref{eq:objective} for the target tracking problem we study
forms a sum over information sources, the targets \eqref{eq:objective_sum}.
Let $\setfuns = \{\setfun_1,\ldots,\setfun_\numtargets\}$
be a collection of set functions so that for $j\in\targets$ and
$X\subseteq\ground$
\begin{align}
  \setfun_j(X) =
  \MI(\targetstates_{j,t+1:t+l};
  \observations_{j,t+1:t+l}(X) |\observations_{j,0:t},\robotstates_{0:t}).
  \label{eq:target_sum_decomposition}
\end{align}
This collection of set functions $\setfuns$ decomposes
$\setfun$
and enables us to characterize interactions between
robots in terms of robots' capacities to sense near and distant targets.
\begin{definition}[Sum decomposition]
  \label{def:sum_submodular}
  A set of submodular, monotonic, normalized functions
  $\setfuns=\{\setfun_1,\ldots,\setfun_n\}$
  decomposes a set function $\setfun$ if
  \begin{align}
    \setfun(X) = \sum_{\hat \setfun \in \setfuns} \hat \setfun(X),
    \quad
    \text{for all } X\subseteq\ground.
    \label{eq:sum_submodular}
  \end{align}
\end{definition}

Closure over sums~\citep{foldes2005} ensures that $\setfun$ is
submodular, monotonic, and normalized if the same is true for each
$\hat\setfun\in\setfuns$.
Further, although some such sum decomposition always exists
($\setfuns\!=\!\{\setfun\}$), the choice of decomposition affects the
tightness of the performance bound; choosing $\setfuns$ to form a sum over
targets \eqref{eq:target_sum_decomposition} will capture spatial locality in
distributions of robots and targets.

\subsubsection{Derivatives and the sum decomposition}
Given some decomposition $\setfuns$ of $\setfun$, the second
derivative~\eqref{eq:derivative} of $\setfun$
at $X\subseteq\ground$ with respect to
$A,B\subseteq\ground$, all disjoint, is%
\begin{align}
  \setfun(A;B|X) = \sum_{\hat \setfun \in \setfuns} \hat \setfun(A;B|X).
  \label{eq:sum_derivative}
\end{align}

This derivative has the form of the negation of the cost of ignoring
decisions during distributed planning
$\distcost$~\eqref{eq:distributed_planning_cost}, and the rest of this section
is devoted to obtaining a bound on expressions which relate a robot's
decision $A$ to the decisions that robot ignores $B$ while eliminating
dependency on which prior decisions the robot has access to $X$.

\subsubsection{Bounding second derivatives via sum decompositions}
Applying monotonicity and submodularity respectively provides a lower
bound on the second derivative of a set function
\begin{align}
  \setfun(A;B|X) = \setfun(A|B,X) - \setfun(A|X) \geq -\setfun(A|X) \geq
  -\setfun(A)
  \label{eq:incremental_bound}
\end{align}
where $A,B,X\subseteq\ground$ are disjoint.
By symmetry
\begin{align}
  \setfun(A;B|X) \geq -\min(\setfun(A),\setfun(B)).
  \label{eq:min_bound}
\end{align}
Then, expressing the second derivative of $\setfun$ in terms of the sum
decomposition~\eqref{eq:sum_derivative} and using \eqref{eq:min_bound} to  bound
the derivatives of $\hat\setfun\in\setfuns$ yields
\begin{align}
  \setfun(A;B|X) \geq \sum_{\hat \setfun\in\setfuns} -\min(\hat \setfun(A),\hat
  \setfun (B)).
  \label{eq:derivative_bound}
\end{align}

\begin{remark}
  Our prior work~\citep{corah2018cdc} relies on $\setfun(A;B|X)$
  increasing monotonically in $X$ to state $\setfun(A;B|X)\geq\setfun(A;B)$.
  To unify this result with \eqref{eq:derivative_bound}, we state that
  each produces a lower bound on $\setfun(A;B|X)$ as a function of $A$ and $B$.
  \label{remark:monotonic_upper_bound}
\end{remark}

\subsubsection{Quantifying inter-robot redundancy}
This bound on the second derivative of $\setfun$ \eqref{eq:derivative_bound}
leads to a bound on redundancy between agents which we express with the weights:
\begin{align}
  \begin{split}
    \weights(i,j)
    &=
    \max_{x_i\in\block_i, x_j\in\block_j}
    \sum_{\hat \setfun\in\setfuns} \min(\hat \setfun(x_i),\hat \setfun (x_j))
    \\
    &\geq -\setfun(x'_i; x'_j | X)
    \\
    &
    \quad \text{for all }
    x'_i\!\in\!\block_i,\ x'_j\!\in\!\block_j,\
    X\subseteq \Omega\!\setminus\!\{x'_i , x'_j\}.
  \end{split}
  \label{eq:weights}
\end{align}
Evaluating values of $\weights$ is difficult as doing so
involves search over the product of two robots' action spaces.
To make evaluation of weights tractable, relaxing this expression by taking
the pairwise minimum of the maximum values of each objective component
produces an upper bound in terms of channel capacities
\eqref{eq:target_capacities}, avoiding search over a product space
\begin{align}
  \begin{split}
    \approxweights(i,j)
    &=
    \sum_{k \in \targets}
    \min\left(
      C_{i,k},
      C_{j,k}
    \right) \\
    &=
    \sum_{\hat \setfun\in\setfuns}
    \min\left(
      \max_{x_i\in\block_i}\hat \setfun(x_i),
      \max_{x_j\in\block_j}\hat \setfun(x_j)
    \right) \\
    &\geq
    \max_{x_i\in\block_i, x_j\in\block_j}
    \sum_{\hat \setfun\in\setfuns} \min(\hat \setfun(x_i),\hat \setfun (x_j))
    =\weights(i,j).
  \end{split}
  \label{eq:weight_by_component}
\end{align}
The second equality follows from the definition of the channel capacities
\eqref{eq:target_capacities},
recalling that 
we chose $\setfuns$ to decompose $\setfun$ by targets $\targets$.
This expresses how \emph{if sensing capacity decreases with distance so
do interactions between robots.}


\subsection{Summary of the proof of Theorem~\ref{theorem:main_bound}}
\label{sec:summary_of_proof_of_main_bound}

Theorem~\ref{theorem:main_bound} consists of two parts.
The first, the effect of approximations on planning performance
\eqref{eq:costs_bound} follows by applying
Lemma~\ref{lemma:approximate_suboptimality}, on the suboptimality of general
assignments, and substituting the definitions of the costs
(\eqref{eq:distributed_planning_cost}, \eqref{eq:objective_cost}, and
\eqref{eq:planner_cost}).
The second part \eqref{eq:weights_bound} characterizes suboptimality due to
distributed planning and follows by applying a chain rule
\ifextended
(Appendix~\ref{appendix:chain_rule})
\else
(Appendix~I)
\fi
to the definition of
cost of distributed planning \eqref{eq:distributed_planning_cost} and
substituting \eqref{eq:derivative_bound}, \eqref{eq:weights}, and
\eqref{eq:weight_by_component}.
Please refer to
\ifextended
Appendix~\ref{sec:proof_of_main_bound} for the full
proof.
\else
Appendix~III for the full proof~\citep{corah2021irosarxiv}.
\fi

\section{Run time and scaling}
\label{sec:scaling}

Algorithm~\ref{alg:distributed} requires a number of sequential planning steps
that depends on the planner graph
(Sec.~\ref{sec:graph_model}), a constant number of steps for \rsp{}.
Further, Sec.~\ref{sec:rsp} stated that
if the optimum is proportional to the sum of weights \eqref{eq:sum_of_weights}
$\rsp$ guarantees suboptimality approaching half of optimal.

Assuming the optimum is proportional to the number of robots ($\numrobots$), then
\eqref{eq:sum_of_weights} holds if the sum of weights is proportional to
$\numrobots$ as well.
\ifextended
Appendix~\ref{sec:scaling_analysis}
\else
Appendix~IV
\fi
presents sufficient conditions for
the sum of weights to be proportional to $\numrobots$,
given the robot-target channel capacities \eqref{eq:target_capacities} are
bounded appropriately with distance.
This ensures the cost of distributed
planning $\distcost$~\eqref{eq:distributed_planning_cost} is bounded
independent of $\numrobots$.

Additionally, bounds on channel capacities as a function of distance
enable robots to ignore sufficiently distant robots and targets
in what we refer to as Range-limited \rsp{} (\rrsp{})~\citep{corah2020phd}
as the additional cost (ignoring decisions and approximating the objective)
approaches zero.

By incorporating range limits and tracking targets with sparse
Bayes filters we also achieve single-robot planning in constant time
(depending on densities of robots and targets).

Regarding communication, robots send one message with constant size for each
edge in the directed planner graph (Sec.~\ref{sec:graph_model}).
Ignoring distant robots via \rrsp{} reduces the total to a constant number of
messages per robot~\citep{corah2018cdc}.

\section{Results}
\label{sec:results}

\begin{figure}
  \includegraphics[width=0.49\linewidth]{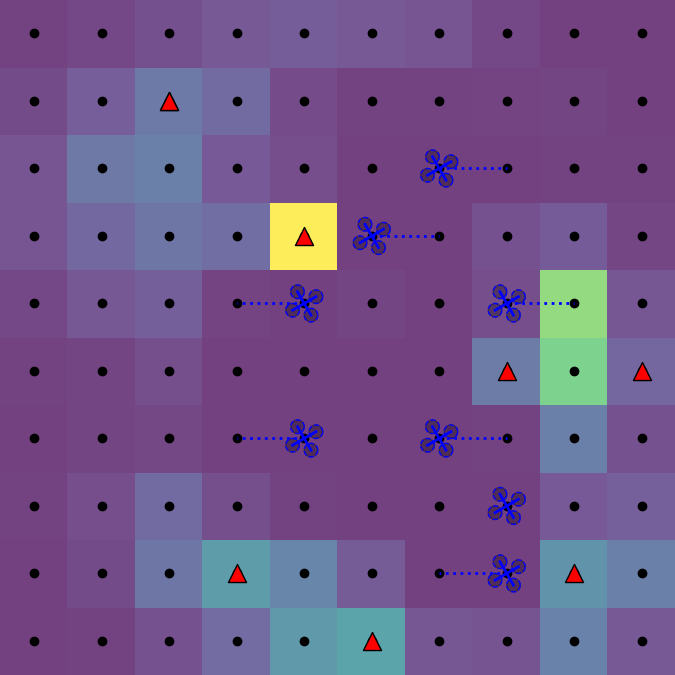}
  \includegraphics[width=0.49\linewidth]{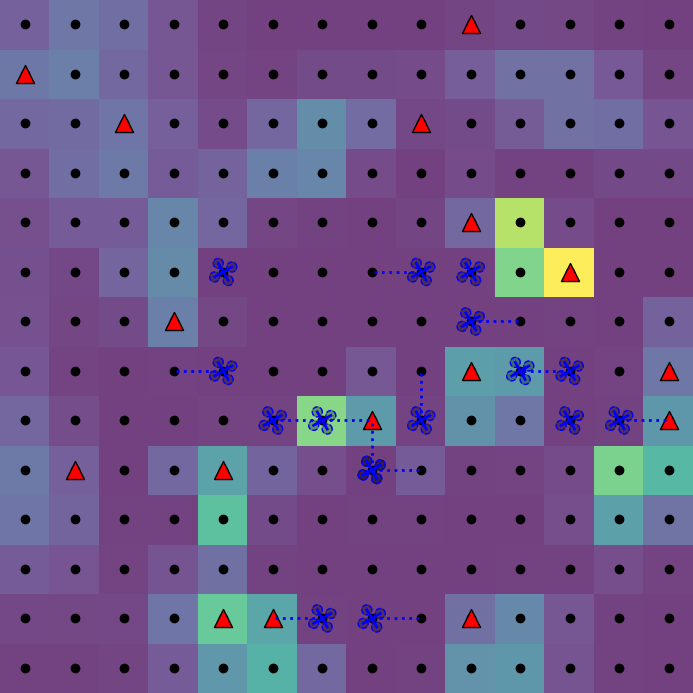}
  \caption{Visualizations of eight and sixteen robots tracking
    same numbers of targets.
    Robots with dotted finite-horizon trajectories are blue and
    targets red.
    The background illustrates the sum of target probabilities at each grid
    space, increasing from purple to yellow.
  }\label{fig:target_tracking_visualization}
\end{figure}

To evaluate the approach, we provide simulation results
(visualized in Fig.~\ref{fig:target_tracking_visualization})
for teams of robots tracking groups of targets (one target per robot).
Robots move according to planner output and targets via a random walk,
all on a square four-connected grid with $\sqrt{12.5\numrobots}$ cells on each
side.\footnote{
  Numbers of targets and grid cells are proportional to the number of
  robots ($\numrobots$).
  We desire entropies approaching a constant
  for large $\numrobots$ on a per-robot basis and the same for redundancy and
  objective values.
}
The robots estimate target locations via Bayesian filters\footnote{
  Additionally, planners with sixteen or more robots use sparse filters,
  ignoring target occupancy probabilities below $10^{-3}$.
}
given range
observations to each target with mean $\hat d = \min(d, 20)$ and variance
$0.25+0.5 \hat d^2$ where $d$ is the Euclidean distance to the target in cells
lengths.\footnote{
  We selected simulation parameters to maximize discrepancy between
  sequential and myopic planning without evaluating the performance of \rsp{}.
}
For the purpose of this paper,
robots have access to all observations or,
equivalently, centralized filters.
Trials run for 100 time-steps; initial states are uniformly
random; initial target positions are known;\footnote{%
  Known initial states promote fast convergence and ensure initial
  uncertainty does not increase with environment size.
}
and we ignore the first 20 steps of each trial to allow the system to converge
to steady-state conditions.

Robots plan actions individually using Monte Carlo tree search
(MCTS, \method{PlanAnytime}) \citep{chaslot2010} with a two step horizon and
collectively according to the specified distributed planner.
To ensure tractability we replace the original objective
\eqref{eq:objective} with a sum of mutual information for each time-step
\begin{align}
  \setfunsim(X) &=
  \sum_{k=1}^l \MI(\targetstates_{t+k}; \observations_{t+1:t+k}(X)
  |\observations_{0:t},\robotstates_{0:t}),\ X\subseteq\ground.
  \label{eq:simulation_objective}
\end{align}
This objective
is equivalent to \citep[(18)]{ryan2010ras}
and
can be thought of as minimizing uncertainty at the time of each
planning step.
Being a sum, \eqref{eq:simulation_objective} remains submodular, monotonic, and
normalized, and all analysis, including Theorem~\ref{theorem:main_bound},
applies unchanged.
Like \citet{ryan2010ras}, we evaluate this objective by simulating the
system and computing the sample mean of filter entropy.
The MCTS planner also estimates the objective
implicitly by simulating the system once per rollout; by sampling more
valuable actions more often, MCTS produces increasingly accurate estimates for
nearly optimal trajectories.
The experiments compare methods for multi-robot coordination including:
sequential planning \eqref{eq:sequential_greedy};
distributed planning (Alg.~\ref{alg:distributed}) with
\rsp{}~\citep{corah2018cdc};
myopic planning (MCTS without coordination);
and random actions.
Additionally, we provide results with range limits (\rrsp{})
where robots ignore targets further than 12 units (in
terms of the filter mean positions) and robots further than 20 units.
Given use of sparse filters, \rrsp{} runs in \emph{constant time}.

\begin{figure}
  \setlength{\subfiguresheight}{1.75in}
  \begin{subfigure}[b]{0.5\linewidth}
    \setlength{\figurewidth}{1.1\linewidth}
    \setlength{\figureheight}{\subfiguresheight}
    \centering
    \tiny
    %
\begin{tikzpicture}

\definecolor{color0}{rgb}{0.12156862745098,0.466666666666667,0.705882352941177}
\definecolor{color1}{rgb}{1,0.498039215686275,0.0549019607843137}
\definecolor{color2}{rgb}{0.172549019607843,0.627450980392157,0.172549019607843}

\begin{axis}[
height=\figureheight,
legend cell align={left},
legend columns=4,
legend style={anchor=south east, at={(1.0, 1.05)}, fill opacity=0.8, transpose legend, draw opacity=1, text opacity=1, draw=white!80.0!black},
legend image post style={scale=0.5}, 
tick align=inside,
tick pos=left,
width=\figurewidth,
x grid style={white!69.01960784313725!black},
xlabel={Number of robots},
xlabel near ticks,
xtick={8,16,24,32,40},
xmajorgrids,
xmin=6.4, xmax=41.6,
xtick style={color=black},
y grid style={white!69.0196078431373!black},
ylabel={Entropy per target (bits)},
ylabel near ticks,
ymajorgrids,
ymin=2.24164019856608, ymax=2.89018918373825,
ytick style={color=black}
]
\path [draw=black, fill=black, opacity=0.2, line width=0pt]
(axis cs:8,2.56954467114637)
--(axis cs:16,2.42235708699605)
--(axis cs:24,2.37695865394502)
--(axis cs:32,2.41260254888404)
--(axis cs:40,2.35518116489349)
--(axis cs:40,2.3257583091782)
--(axis cs:32,2.38978170228155)
--(axis cs:24,2.34551408461984)
--(axis cs:16,2.39292492502629)
--(axis cs:8,2.54288318682618)
--cycle;
\path [draw=black, fill=black, opacity=0.2, line width=0pt, dashed]
(axis cs:8,2.53233788649251)
--(axis cs:16,2.38812120998565)
--(axis cs:24,2.3226670843118)
--(axis cs:32,2.36055831630139)
--(axis cs:40,2.31823686051357)
--(axis cs:40,2.29670758162895)
--(axis cs:32,2.32664331553059)
--(axis cs:24,2.28407605000135)
--(axis cs:16,2.35400746946649)
--(axis cs:8,2.48630423737665)
--cycle;
\path [draw=black, fill=black, opacity=0.2, line width=0pt, dotted]
(axis cs:8,2.55014514705743)
--(axis cs:16,2.36970048142801)
--(axis cs:24,2.31323636459936)
--(axis cs:32,2.34425678224304)
--(axis cs:40,2.30193306885011)
--(axis cs:40,2.28209480982017)
--(axis cs:32,2.30071406069407)
--(axis cs:24,2.27611636350348)
--(axis cs:16,2.34041889202868)
--(axis cs:8,2.51732737389858)
--cycle;
\path [draw=black, fill=black, opacity=0.2, line width=0pt, dash pattern=on 1pt off 3pt on 3pt off 3pt]
(axis cs:8,2.57169183030332)
--(axis cs:16,2.36572169416829)
--(axis cs:24,2.323593881522)
--(axis cs:32,2.36274578182163)
--(axis cs:40,2.29319176446206)
--(axis cs:40,2.27111969789209)
--(axis cs:32,2.33250533568819)
--(axis cs:24,2.28869402391063)
--(axis cs:16,2.33854933644551)
--(axis cs:8,2.54528974041867)
--cycle;
\path [draw=color0, fill=color0, opacity=0.2, line width=0pt]
(axis cs:8,2.86070968441224)
--(axis cs:16,2.69856155830629)
--(axis cs:24,2.61311578728931)
--(axis cs:32,2.63945381833043)
--(axis cs:40,2.56277057968117)
--(axis cs:40,2.53337411407392)
--(axis cs:32,2.60497126906484)
--(axis cs:24,2.5865314785687)
--(axis cs:16,2.65719882985572)
--(axis cs:8,2.82976826205221)
--cycle;
\path [draw=color1, fill=color1, opacity=0.2, line width=0pt]
(axis cs:8,2.69201233792399)
--(axis cs:16,2.61454500221667)
--(axis cs:24,2.59470503540719)
--(axis cs:32,2.67659238329188)
--(axis cs:40,2.66314663867345)
--(axis cs:40,2.60106394223979)
--(axis cs:32,2.63252919865644)
--(axis cs:24,2.55507647673788)
--(axis cs:16,2.57859460844787)
--(axis cs:8,2.67238046889369)
--cycle;
\path [draw=color2, fill=color2, opacity=0.2, line width=0pt]
(axis cs:8,2.56749689181074)
--(axis cs:16,2.35228653080604)
--(axis cs:24,2.31505384360627)
--(axis cs:32,2.35047621509369)
--(axis cs:40,2.30927474837588)
--(axis cs:40,2.28246994412093)
--(axis cs:32,2.3229881308887)
--(axis cs:24,2.28009900753928)
--(axis cs:16,2.31457519071448)
--(axis cs:8,2.53046199303333)
--cycle;
\addplot [semithick, black]
table {%
8 2.55621392898628
16 2.40764100601117
24 2.36123636928243
32 2.40119212558279
40 2.34046973703584
};
\addlegendentry{\rsp[][\textbf]{} $\bm{\numrounds\!=\!2}$}
\addplot [semithick, black, dashed]
table {%
8 2.50932106193458
16 2.37106433972607
24 2.30337156715657
32 2.34360081591599
40 2.30747222107126
};
\addlegendentry{\rsp[][\textbf]{} $\bm{\numrounds\!=\!4}$}
\addplot [semithick, black, dotted]
table {%
8 2.53373626047801
16 2.35505968672835
24 2.29467636405142
32 2.32248542146856
40 2.29201393933514
};
\addlegendentry{\rsp[][\textbf]{} $\bm{\numrounds\!=\!8}$}
\addplot [semithick, black, dash pattern=on 1pt off 3pt on 3pt off 3pt]
table {%
8 2.55849078536099
16 2.3521355153069
24 2.30614395271632
32 2.34762555875491
40 2.28215573117707
};
\addlegendentry{\rrsp[][\textbf]{} $\bm{\numrounds\!=\!4}$}
\addplot [semithick, color0]
table {%
8 2.84523897323223
16 2.677880194081
24 2.59982363292901
32 2.62221254369764
40 2.54807234687755
};
\addlegendentry{random}
\addplot [semithick, color1]
table {%
8 2.68219640340884
16 2.59656980533227
24 2.57489075607254
32 2.65456079097416
40 2.63210529045662
};
\addlegendentry{myopic}
\addplot [semithick, color2]
table {%
8 2.54897944242204
16 2.33343086076026
24 2.29757642557278
32 2.3367321729912
40 2.2958723462484
};
\addlegendentry{sequential}
\end{axis}

\end{tikzpicture}%

    \caption{
      Target entropy
    }\label{subfig:target_entropy}
  \end{subfigure}%
  \begin{subfigure}[b]{0.5\linewidth}
    \setlength{\figurewidth}{1.1\linewidth}%
    \setlength{\figureheight}{1.26\subfiguresheight}%
    \centering%
    \tiny%
    \hskip-0.45cm%
    %
\begin{tikzpicture}

\definecolor{color0}{rgb}{1,0.498039215686275,0.0549019607843137}

\begin{axis}[
height=\figureheight,
tick align=outside,
tick pos=left,
width=\figurewidth,
x grid style={white!69.01960784313725!black},
xlabel={Objective (fraction of max)},
xmin=0.878939166426098, xmax=1.00576480159876,
xtick style={color=black},
y grid style={white!69.01960784313725!black},
ymin=0.5, ymax=7.5,
ytick style={color=black},
ytick={1,2,3,4,5,6,7},
yticklabel style={rotate=45,anchor=east},
yticklabels={
random,
myopic,
\rsp[][\textbf]{} $\bm{\numrounds\!=\!2}$,
\rsp[][\textbf]{} $\bm{\numrounds\!=\!4}$,
\rsp[][\textbf]{} $\bm{\numrounds\!=\!8}$,
\rrsp[][\textbf]{} $\bm{\numrounds\!=\!4}$,
sequential
}
]
\addplot [black]
table {%
0.954787286286957 0.75
0.954787286286957 1.25
0.979055059400186 1.25
0.979055059400186 0.75
0.954787286286957 0.75
};
\addplot [black]
table {%
0.954787286286957 1
0.918680839191225 1
};
\addplot [black]
table {%
0.979055059400186 1
1 1
};
\addplot [black]
table {%
0.918680839191225 0.875
0.918680839191225 1.125
};
\addplot [black]
table {%
1 0.875
1 1.125
};
\addplot [black, mark=*, mark size=2, mark options={solid,fill opacity=0}, only marks]
table {%
0.918280288978132 1
0.918108995934431 1
0.91529252463204 1
0.890974614023062 1
0.912969454920479 1
0.916320355865791 1
0.890636006826516 1
0.906709209038523 1
0.915273640836571 1
0.884703968024856 1
0.904413518439951 1
0.909402495180871 1
0.912327675975411 1
0.909012620589598 1
0.917763084593916 1
0.910079068408362 1
0.908133155977623 1
0.916971106133061 1
0.91581185724962 1
0.918206123669863 1
0.90992095701077 1
0.894127994044857 1
0.91190662651904 1
0.892428996293456 1
0.916501431488103 1
0.916510658669993 1
0.917252434782623 1
};
\addplot [black]
table {%
0.965165629163503 1.75
0.965165629163503 2.25
0.989857796676087 2.25
0.989857796676087 1.75
0.965165629163503 1.75
};
\addplot [black]
table {%
0.965165629163503 2
0.928340231907027 2
};
\addplot [black]
table {%
0.989857796676087 2
1 2
};
\addplot [black]
table {%
0.928340231907027 1.875
0.928340231907027 2.125
};
\addplot [black]
table {%
1 1.875
1 2.125
};
\addplot [black, mark=*, mark size=2, mark options={solid,fill opacity=0}, only marks]
table {%
0.893429786100883 2
0.919638233237414 2
0.913503824204114 2
0.920749111720918 2
0.927699921512262 2
0.920881834906844 2
0.912112739981062 2
0.909271118577496 2
0.927812744207486 2
0.922146411825504 2
0.920698819235966 2
0.919977779902339 2
0.919289349916075 2
0.898340783179206 2
0.927551717636485 2
0.918322469353524 2
0.925802491300633 2
0.926343718077477 2
0.919437244020632 2
0.922806780259107 2
0.910174561011996 2
0.924458192644779 2
0.919512711117798 2
0.925449603657748 2
0.928002959225308 2
0.927503589261395 2
};
\addplot [black]
table {%
0.980248303792678 2.75
0.980248303792678 3.25
0.995752253704548 3.25
0.995752253704548 2.75
0.980248303792678 2.75
};
\addplot [black]
table {%
0.980248303792678 3
0.957050396478208 3
};
\addplot [black]
table {%
0.995752253704548 3
1 3
};
\addplot [black]
table {%
0.957050396478208 2.875
0.957050396478208 3.125
};
\addplot [black]
table {%
1 2.875
1 3.125
};
\addplot [black, mark=*, mark size=2, mark options={solid,fill opacity=0}, only marks]
table {%
0.953286822395311 3
0.945682932480194 3
0.956746583408294 3
0.946542851435617 3
0.951879853610367 3
0.954109042682895 3
0.953386755461595 3
0.953695680664574 3
0.955744953099545 3
0.951378058301163 3
0.954476271015511 3
0.95248994258515 3
0.953017825990497 3
0.937861736111546 3
0.954155406826504 3
0.9529908276563 3
0.940237759936091 3
0.946909794740719 3
0.950934627859721 3
};
\addplot [black]
table {%
0.981943678740974 3.75
0.981943678740974 4.25
0.997291526780195 4.25
0.997291526780195 3.75
0.981943678740974 3.75
};
\addplot [black]
table {%
0.981943678740974 4
0.959232427628575 4
};
\addplot [black]
table {%
0.997291526780195 4
1 4
};
\addplot [black]
table {%
0.959232427628575 3.875
0.959232427628575 4.125
};
\addplot [black]
table {%
1 3.875
1 4.125
};
\addplot [black, mark=*, mark size=2, mark options={solid,fill opacity=0}, only marks]
table {%
0.953321150917319 4
0.957003207618879 4
0.954985896136685 4
0.949939233472981 4
0.943823726014113 4
0.950189552724365 4
0.956177656850262 4
0.949979566489632 4
0.958181486927296 4
0.958310347435815 4
0.957559327357968 4
0.956610392930365 4
0.957063531383561 4
0.952814988615422 4
0.952634737753976 4
0.951334097453654 4
0.957440141043636 4
0.94209789584666 4
};
\addplot [black]
table {%
0.981928887909064 4.75
0.981928887909064 5.25
0.997484887488122 5.25
0.997484887488122 4.75
0.981928887909064 4.75
};
\addplot [black]
table {%
0.981928887909064 5
0.958610849657073 5
};
\addplot [black]
table {%
0.997484887488122 5
1 5
};
\addplot [black]
table {%
0.958610849657073 4.875
0.958610849657073 5.125
};
\addplot [black]
table {%
1 4.875
1 5.125
};
\addplot [black, mark=*, mark size=2, mark options={solid,fill opacity=0}, only marks]
table {%
0.954806118116995 5
0.95267106533295 5
0.953943714154223 5
0.949585034863671 5
0.947322139486869 5
0.941890497796306 5
0.94815758374126 5
0.952339181083388 5
0.952710450954984 5
0.949714515017142 5
0.939975637798977 5
0.955423484103426 5
0.954494096361914 5
0.946080938006907 5
0.943901666044695 5
0.949414375651824 5
0.954476598208151 5
0.957052998308692 5
0.952431394078387 5
0.958182176545859 5
0.942718548910982 5
0.954769229481822 5
};
\addplot [black]
table {%
0.981653037132224 5.75
0.981653037132224 6.25
0.997559681880982 6.25
0.997559681880982 5.75
0.981653037132224 5.75
};
\addplot [black]
table {%
0.981653037132224 6
0.957906902600171 6
};
\addplot [black]
table {%
0.997559681880982 6
1 6
};
\addplot [black]
table {%
0.957906902600171 5.875
0.957906902600171 6.125
};
\addplot [black]
table {%
1 5.875
1 6.125
};
\addplot [black, mark=*, mark size=2, mark options={solid,fill opacity=0}, only marks]
table {%
0.955806619653948 6
0.951215153276078 6
0.951754739792571 6
0.95473215597452 6
0.955310163365054 6
0.949406450272575 6
0.955574033697549 6
0.957338494356866 6
0.954552701303264 6
0.955028596084088 6
0.957057592274267 6
0.948264289017008 6
0.942818541865282 6
0.954669810744472 6
};
\addplot [black]
table {%
0.981593215305093 6.75
0.981593215305093 7.25
0.998463263398228 7.25
0.998463263398228 6.75
0.981593215305093 6.75
};
\addplot [black]
table {%
0.981593215305093 7
0.957007125404814 7
};
\addplot [black]
table {%
0.998463263398228 7
1 7
};
\addplot [black]
table {%
0.957007125404814 6.875
0.957007125404814 7.125
};
\addplot [black]
table {%
1 6.875
1 7.125
};
\addplot [black, mark=*, mark size=2, mark options={solid,fill opacity=0}, only marks]
table {%
0.95164684295425 7
0.953132083503165 7
0.94655618691 7
0.951041402533606 7
0.950621936571106 7
0.949416165492488 7
0.955839430594991 7
0.955238808562277 7
0.953376121194091 7
0.95191119242797 7
0.955353189680017 7
0.955280215907544 7
0.954083070155499 7
0.955207187430723 7
0.951006789289451 7
0.955902276689458 7
0.943251722259929 7
0.947917430557254 7
0.948435703628967 7
0.942473068539862 7
};
\addplot [color0]
table {%
0.968063591025807 0.75
0.968063591025807 1.25
};
\addplot [color0]
table {%
0.97826669395923 1.75
0.97826669395923 2.25
};
\addplot [color0]
table {%
0.98832324562467 2.75
0.98832324562467 3.25
};
\addplot [color0]
table {%
0.989810920883941 3.75
0.989810920883941 4.25
};
\addplot [color0]
table {%
0.990164835412537 4.75
0.990164835412537 5.25
};
\addplot [color0]
table {%
0.99046737965763 5.75
0.99046737965763 6.25
};
\addplot [color0]
table {%
0.990082800270504 6.75
0.990082800270504 7.25
};
\end{axis}

\end{tikzpicture}%
%
    \caption{
      Objective values
    }\label{subfig:solver_objectives}
  \end{subfigure}
  \begin{subfigure}[b]{\linewidth}
    \setlength{\figurewidth}{1.0\linewidth}
    \setlength{\figureheight}{1.0\subfiguresheight}
    \centering
    \tiny
    %
\begin{tikzpicture}

\definecolor{color0}{rgb}{0.12156862745098,0.466666666666667,0.705882352941177}
\definecolor{color1}{rgb}{1,0.498039215686275,0.0549019607843137}

\pgfplotsset{set layers}
\begin{axis}[
height=\figureheight,
tick align=inside,
tick pos=left,
width=\figurewidth,
x grid style={white!69.0196078431373!black},
xmin=3.6, xmax=100.4,
ytick={10,20,30,40,50},
xtick style={color=black},
y grid style={white!69.0196078431373!black},
ylabel={Redundancy per robot},
ymin=8.02853323347533, ymax=59.1780966338163,
ytick style={color=black}
]
\path [draw=color0, fill=color0, opacity=0.2, line width=0pt]
(axis cs:8,11.3278873085606)
--(axis cs:16,25.9376675444271)
--(axis cs:24,39.6098885870783)
--(axis cs:32,45.3601595597765)
--(axis cs:40,38.402234978364)
--(axis cs:48,56.8531164792553)
--(axis cs:56,49.8695555712915)
--(axis cs:64,39.4377740457706)
--(axis cs:72,37.0149390847419)
--(axis cs:80,36.2801596696439)
--(axis cs:88,37.9826623058646)
--(axis cs:96,28.6300737274411)
--(axis cs:96,24.7707291005357)
--(axis cs:88,33.5162977952598)
--(axis cs:80,30.0056346530759)
--(axis cs:72,28.6496842810357)
--(axis cs:64,32.2086825616451)
--(axis cs:56,42.155327320621)
--(axis cs:48,39.1618150120418)
--(axis cs:40,33.8463739563353)
--(axis cs:32,38.1857786313987)
--(axis cs:24,32.534079685877)
--(axis cs:16,23.2434059225344)
--(axis cs:8,10.3535133880363)
--cycle;
\path [draw=color0, fill=color0, opacity=0.2, line width=0pt]
(axis cs:8,12.0825507106068)
--(axis cs:16,25.0749436426598)
--(axis cs:24,36.3811103113742)
--(axis cs:32,38.7860552808557)
--(axis cs:40,43.738337275971)
--(axis cs:40,39.0474713311373)
--(axis cs:32,35.6934229443658)
--(axis cs:24,33.5734975028841)
--(axis cs:16,23.5062171293475)
--(axis cs:8,11.3948591113165)
--cycle;
\addplot [very thick, color0, mark=*, mark size=1, mark options={solid}]
table {%
8 10.8407003482985
16 24.5905367334807
24 36.0719841364776
32 41.7729690955876
40 36.1243044673497
48 48.0074657456485
56 46.0124414459563
64 35.8232283037079
72 32.8323116828888
80 33.1428971613599
88 35.7494800505622
96 26.7004014139884
};
\label{plot:one}
\addplot [very thick, color0, dashed, mark=*, mark size=1, mark options={solid}]
table {%
8 11.7387049109616
16 24.2905803860036
24 34.9773039071291
32 37.2397391126107
40 41.3929043035541
};
\label{plot:two}
\end{axis}

\begin{axis}[
axis y line*=right, 
height=\figureheight,
tick align=inside,
legend cell align={left},
legend columns=2,
legend style={
  fill opacity=0.8,
  transpose legend,
  draw opacity=1,
  text opacity=1,
  at={(0.97,0.03)},
  anchor=south east,
  draw=white!80!black
},
width=\figurewidth,
x grid style={white!69.0196078431373!black},
xlabel={Number of robots},
xmin=3.6, xmax=100.4,
xtick pos=left,
xtick style={color=black},
y grid style={white!69.0196078431373!black},
ylabel={Objective value per robot},
ymin=2.39466482293086, ymax=3.23828019543246,
ytick pos=right,
ytick style={color=black},
ytick={2.4,2.5,2.6,2.7,2.8,2.9,3,3.1,3.2,3.3},
yticklabel style={anchor=west},
]
\addlegendimage{/pgfplots/refstyle=plot:one}
\addlegendentry{redundancy, large-scale}
\addlegendimage{/pgfplots/refstyle=plot:two}
\addlegendentry{redundancy, solver-trials}

\path [draw=color1, fill=color1, opacity=0.2, line width=0pt]
(axis cs:8,2.5027583287903)
--(axis cs:16,2.84246688930201)
--(axis cs:24,3.00138344344998)
--(axis cs:32,3.08448309541823)
--(axis cs:40,3.04614188348781)
--(axis cs:48,3.18057835310664)
--(axis cs:56,3.18207617522624)
--(axis cs:64,3.14336207640371)
--(axis cs:72,3.12554849702344)
--(axis cs:80,3.14304665240979)
--(axis cs:88,3.19993404213693)
--(axis cs:96,3.13283244184965)
--(axis cs:96,3.09878007634196)
--(axis cs:88,3.13791903295212)
--(axis cs:80,3.08796922227916)
--(axis cs:72,3.06316337694661)
--(axis cs:64,3.06682820699135)
--(axis cs:56,3.10550999448357)
--(axis cs:48,3.10717151121534)
--(axis cs:40,2.99655517449246)
--(axis cs:32,3.03228295055599)
--(axis cs:24,2.90169412712356)
--(axis cs:16,2.77379766408576)
--(axis cs:8,2.43301097622638)
--cycle;
\path [draw=color1, fill=color1, opacity=0.2, line width=0pt]
(axis cs:8,2.61097159976645)
--(axis cs:16,2.84314914782696)
--(axis cs:24,2.99093438996726)
--(axis cs:32,2.97774441413539)
--(axis cs:40,3.03899626478092)
--(axis cs:40,3.00103934361677)
--(axis cs:32,2.93086470237687)
--(axis cs:24,2.93589053013479)
--(axis cs:16,2.78992901163479)
--(axis cs:8,2.54900118867097)
--cycle;
\addplot [very thick, color1, mark=*, mark size=1, mark options={solid}]
table {%
8 2.46788465250834
16 2.80813227669389
24 2.95153878528677
32 3.05838302298711
40 3.02134852899013
48 3.14387493216099
56 3.1437930848549
64 3.10509514169753
72 3.09435593698502
80 3.11550793734448
88 3.16892653754453
96 3.1158062590958
};
\addlegendentry{objective, large-scale}
\addplot [very thick, color1, dashed, mark=*, mark size=1, mark options={solid}]
table {%
8 2.57998639421871
16 2.81653907973087
24 2.96341246005103
32 2.95430455825613
40 3.02001780419884
};
\addlegendentry{objective, solver-trials}
\end{axis}

\end{tikzpicture}%

    \caption{
      Scaling of objective and redundancy~\eqref{eq:sum_of_weights}
    }\label{subfig:target_weights}
  \end{subfigure}
  \caption{
    (\subref{subfig:target_entropy})
    Regarding target entropy, the task performance criterion, (lower is better)
    our \rsp{} planners consistently improve upon myopic
    planning and approach sequential planning with many times fewer
    sequential steps.
    (\subref{subfig:solver_objectives})
    Objective values on common 16-robot subproblems reflect a similar trend, and
    results for distributed planning (bold) closely match sequential.
    \eqref{eq:optimization_problem}.
    (\subref{subfig:target_weights})
    Considering average objective values and total
    redundancy~\eqref{eq:sum_of_weights} (both per robot) for \rsp{} with
    $\numrounds\!=\!4$
    (including values for trials from (\subref{subfig:solver_objectives})
    and five additional trials with \rrsp{} with up to 96 robots),
    values initially increase but appear to approach
    asymptotes, indicating that the two
    are approximately proportional at large scales
    \eqref{eq:sum_of_weights}
    so that \rsp{} planning approaches constant-factor suboptimality.
    Shaded regions depict standard error.
  }\label{fig:performance}
\end{figure}
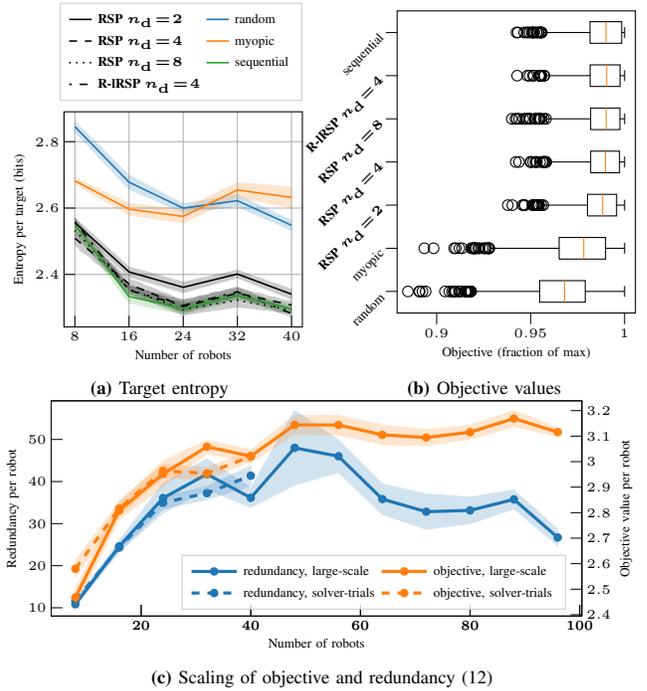

We evaluate the distributed planning approach via
\emph{task performance} (average target entropy) for various numbers of robots
(Fig.~\ref{fig:performance}),
\emph{objective values} on a common set of subproblems
\eqref{eq:optimization_problem},
and the \emph{redundancy per robot}
(the sum of weights \eqref{eq:sum_of_weights} divided by $\numrobots$).
The results for average target entropy---which express uncertainty in
target locations~\citep{cover2012}---are based on 20 simulations of target
tracking for each configuration.
Results for objective values and redundancy use
planning subproblems~\eqref{eq:optimization_problem} taken from the simulation
trials for \rsp{} with $\numrounds\!=\!4$.
The results for objective values by solver are for 16 robots and are normalized
according to maximum values across solvers for each planning problem.
For results on redundancy, an additional five trials of $\rrsp$ with
$\numrounds\!=\!4$ demonstrate behavior for up to 96 robots.\footnote{%
  Planning at this scale is intractable for other planner configurations.
}

Proposed distributed planners provide consistent improvements in target tracking
performance (average target entropy) (Fig.~\ref{subfig:target_entropy}) compared
to myopic planning;
distributed planning in eight rounds matches sequential planning despite
requiring as much as five times fewer planning steps and produces 5--13\% better
(lower) target entropy than when planning myopically.
The objective values (Fig.~\ref{subfig:solver_objectives}) exhibit a similar
trend, and all distributed planners closely match sequential planning.
Although the objective values and redundancy per robot
(Fig.~\ref{subfig:target_weights}) initially increase with more robots, the
values eventually level off and are roughly
proportional~\eqref{eq:sum_of_weights}.
This is consistent with analysis of scaling behavior in
\ifextended
Appendix~\ref{appendix:approximate_suboptimality}
\else
Appendix~II
\fi
which suggests that redundancy per robot approaches a constant value.
Overall, the results indicate that even a small amount of coordination
($\numrounds\!=\!2$),
independent of the number of robots, is sufficient to provide
performance comparable to sequential planning in receding-horizon settings.

\todo{Consider stating more clearly how weights relate to constant-factor
performance}

\section{Conclusions and future work}

This paper has presented a distributed planner for mutual information-based
target tracking which runs in a fixed number of steps and mitigates growth in
planning time for existing sequential planners for submodular maximization.
The analysis provided a novel bound on suboptimality by using target
independence to decompose the objective as a sum.
Additionally, by explicitly accounting for suboptimal local planning (e.g.
anytime planning) and approximation of the objective, we affirmed that the
proposed approach is applicable to practical tracking systems.
The results demonstrated that distributed planning improves tracking
performance (in terms of target entropy) compared to planners with no
coordination and that distributed planning with little coordination can even
match fully sequential planning given a constant number of planning rounds.
Finally, although we focused on target tracking, future work may take advantage
of how the analysis applies to general multi-objective sensing problems,
possibly in concert with our work on coverage~\citep{corah2018cdc}.



\bibliographystyle{IEEEtranN}
{\scriptsize
\balance
\bibliography{bibliography}%

\begin{thebibliography}{37}
\providecommand{\natexlab}[1]{#1}
\providecommand{\url}[1]{#1}
\csname url@samestyle\endcsname
\providecommand{\newblock}{\relax}
\providecommand{\bibinfo}[2]{#2}
\providecommand{\BIBentrySTDinterwordspacing}{\spaceskip=0pt\relax}
\providecommand{\BIBentryALTinterwordstretchfactor}{4}
\providecommand{\BIBentryALTinterwordspacing}{\spaceskip=\fontdimen2\font plus
\BIBentryALTinterwordstretchfactor\fontdimen3\font minus
  \fontdimen4\font\relax}
\providecommand{\BIBforeignlanguage}[2]{{%
\expandafter\ifx\csname l@#1\endcsname\relax
\typeout{** WARNING: IEEEtranN.bst: No hyphenation pattern has been}%
\typeout{** loaded for the language `#1'. Using the pattern for}%
\typeout{** the default language instead.}%
\else
\language=\csname l@#1\endcsname
\fi
#2}}
\providecommand{\BIBdecl}{\relax}
\BIBdecl

\bibitem[Cliff et~al.(2015)Cliff, Fitch, Sukkarieh, Saunders, and
  Heinsohn]{cliff2015rss}
O.~M. Cliff, R.~Fitch, S.~Sukkarieh, D.~L. Saunders, and R.~Heinsohn, ``Online
  localization of radio-tagged wildlife with an autonomous aerial robot
  system,'' in \emph{Proc. of Robot.: Sci. and Syst.}, Rome, Italy, Jul. 2015.

\bibitem[Shah and Schwager(2019)]{shah2019ras}
K.~Shah and M.~Schwager, ``{GRAPE}: Geometric risk-aware pursuit-evasion,''
  \emph{Robot. Auton. Syst.}, 2019.

\bibitem[Charrow et~al.(2014{\natexlab{a}})Charrow, Kumar, and
  Michael]{charrow2014auro}
B.~Charrow, V.~Kumar, and N.~Michael, ``Approximate representations for
  multi-robot control policies that maximize mutual information,'' \emph{Auton.
  Robots}, vol.~37, no.~4, pp. 383--400, 2014.

\bibitem[Charrow et~al.(2014{\natexlab{b}})Charrow, Michael, and
  Kumar]{charrow2014ijrr}
B.~Charrow, N.~Michael, and V.~Kumar, ``Cooperative multi-robot estimation and
  control for radio source localization,'' \emph{Intl. Journal of Robotics
  Research}, vol.~33, no.~4, pp. 569--580, Apr. 2014.

\bibitem[Piacentini et~al.(2019)Piacentini, Bernardini, and
  Beck]{piacentini2019jair}
C.~Piacentini, S.~Bernardini, and J.~C. Beck, ``Autonomous target search with
  multiple coordinated {UAVs},'' \emph{J. Artif. Intell. Research}, vol.~65,
  pp. 519--568, 2019.

\bibitem[Hollinger et~al.(2009)Hollinger, Singh, Djugash, and
  Kehagias]{hollinger2009ijrr}
G.~Hollinger, S.~Singh, J.~Djugash, and A.~Kehagias, ``Efficient multi-robot
  search for a moving target,'' \emph{Intl. Journal of Robotics Research},
  vol.~28, no.~2, pp. 201--219, 2009.

\bibitem[Wang et~al.(2019)Wang, Zhang, Bertinetto, Hu, and Torr]{wang2019cvpr}
Q.~Wang, L.~Zhang, L.~Bertinetto, W.~Hu, and P.~H. Torr, ``Fast online object
  tracking and segmentation: A unifying approach,'' in \emph{Proc. of the
  IEEE/CVF Conf. on Comput. Vis. and Pattern Recognition}, Long Beach, CA, Jun.
  2019.

\bibitem[Chung et~al.(2011)Chung, Hollinger, and Isler]{chung2011auro}
T.~H. Chung, G.~A. Hollinger, and V.~Isler, ``Search and pursuit-evasion in
  mobile robotics,'' \emph{Auton. Robots}, vol.~31, no.~4, p. 299, 2011.

\bibitem[Tokekar et~al.(2014)Tokekar, Isler, and Franchi]{tokekar2014iros}
P.~Tokekar, V.~Isler, and A.~Franchi, ``Multi-target visual tracking with
  aerial robots,'' in \emph{Proc. of the {IEEE/RSJ} Intl. Conf. on Intell.
  Robots and Syst.}, Chicago, IL, Sep. 2014.

\bibitem[Zhou and Tokekar(2019)]{zhou2019tro}
L.~Zhou and P.~Tokekar, ``Sensor assignment algorithms to improve observability
  while tracking targets,'' \emph{{IEEE} Trans. Robotics}, vol.~35, no.~5, pp.
  1206--1219, 2019.

\bibitem[Krause and Guestrin(2005)]{krause2005uai}
A.~Krause and C.~E. Guestrin, ``Near-optimal nonmyopic value of information in
  graphical models,'' in \emph{Proc. of the Conf. on Uncertainty in Artif.
  Intell.}, Edinburgh, Scotland, Jul. 2005.

\bibitem[Schrijver(2003)]{schrijver2003}
A.~Schrijver, \emph{Combinatorial optimization: polyhedra and
  efficiency}.\hskip 1em plus 0.5em minus 0.4em\relax Springer Science \&
  Business Media, 2003, vol.~24.

\bibitem[Singh et~al.(2009)Singh, Krause, Guestrin, and Kaiser]{singh2009}
A.~Singh, A.~Krause, C.~Guestrin, and W.~J. Kaiser, ``Efficient informative
  sensing using multiple robots,'' \emph{J. Artif. Intell. Res.}, vol.~34, pp.
  707--755, 2009.

\bibitem[Feige(1998)]{feige1998}
U.~Feige, ``A threshold of ln $n$ for approximating set cover,'' \emph{Journal
  of the ACM (JACM)}, vol.~45, no.~4, pp. 634--652, 1998.

\bibitem[Atanasov et~al.(2015)Atanasov, {Le Ny}, Daniilidis, and
  Pappas]{atanasov2015icra}
N.~A. Atanasov, J.~{Le Ny}, K.~Daniilidis, and G.~J. Pappas, ``Decentralized
  active information acquisition: Theory and application to multi-robot
  {SLAM},'' in \emph{Proc. of the {IEEE} Intl. Conf. on Robot. and Autom.},
  Seattle, WA, May 2015.

\bibitem[Corah and Michael(2018)]{corah2018cdc}
M.~Corah and N.~Michael, ``Distributed submodular maximization on partition
  matroids for planning on large sensor networks,'' in \emph{Proc. of the
  {IEEE} Conf. on Decision and Control}, Miami, FL, Dec. 2018.

\bibitem[Sung et~al.(2019)Sung, Budhiraja, Williams, and Tokekar]{sung2019auro}
Y.~Sung, A.~K. Budhiraja, R.~K. Williams, and P.~Tokekar, ``Distributed
  assignment with limited communication for multi-robot multi-target
  tracking,'' \emph{Auton. Robots}, 2019.

\bibitem[Corah(2020)]{corah2020phd}
M.~Corah, ``Sensor planning for large numbers of robots,'' Ph.D. dissertation,
  Carnegie Mellon University, 2020.

\bibitem[Fisher et~al.(1978)Fisher, Nemhauser, and Wolsey]{fisher1978}
M.~L. Fisher, G.~L. Nemhauser, and L.~A. Wolsey, ``An analysis of
  approximations for maximizing submodular set functions-{II},''
  \emph{Polyhedral Combinatorics}, vol.~8, pp. 73--87, 1978.

\bibitem[Jorgensen et~al.(2017)Jorgensen, Chen, Milam, and
  Pavone]{jorgensen2017iros}
S.~Jorgensen, R.~H. Chen, M.~B. Milam, and M.~Pavone, ``The matroid team
  surviving orienteers problem: Constrained routing of heterogeneous teams with
  risky traversal,'' in \emph{Proc. of the {IEEE/RSJ} Intl. Conf. on Intell.
  Robots and Syst.}, Vancouver, Canada, Sep. 2017.

\bibitem[Corah and Michael(2019)]{corah2019auro}
M.~Corah and N.~Michael, ``Distributed matroid-constrained submodular
  maximization for multi-robot exploration: theory and practice,'' \emph{Auton.
  Robots}, vol.~43, no.~2, pp. 485--501, 2019.

\bibitem[Chekuri and Martin(2005)]{chekuri2005}
C.~Chekuri and P.~Martin, ``A recursive greedy algorithm for walks in directed
  graphs,'' in \emph{Proc. of the {IEEE} Annu. Symp. Found. Comput. Sci.},
  2005, pp. 245--253.

\bibitem[Hollinger and Sukhatme(2014)]{hollinger2014ijrr}
G.~A. Hollinger and G.~S. Sukhatme, ``Sampling-based robotic information
  gathering algorithms,'' \emph{Intl. Journal of Robotics Research}, vol.~33,
  no.~9, pp. 1271--1287, 2014.

\bibitem[Lauri and Ritala(2016)]{lauri2015ras}
M.~Lauri and R.~Ritala, ``Planning for robotic exploration based on forward
  simulation,'' \emph{Robot. Auton. Syst.}, vol.~83, pp. 15--31, 2016.

\bibitem[Gharesifard and Smith(2017)]{gharesifard2017}
B.~Gharesifard and S.~L. Smith, ``Distributed submodular maximization with
  limited information,'' \emph{{IEEE} Trans. Control Netw. Syst.}, vol.~5,
  no.~4, pp. 1635--1645, 2017.

\bibitem[Grimsman et~al.(2018)Grimsman, Ali, Hespanha, and
  Marden]{grimsman2018tcns}
D.~Grimsman, M.~S. Ali, J.~P. Hespanha, and J.~R. Marden, ``The impact of
  information in greedy submodular maximization,'' \emph{{IEEE} Trans. Control
  Netw. Syst.}, 2018.

\bibitem[Krause et~al.(2008)Krause, Singh, and Guestrin]{krause2008jmlr}
A.~Krause, A.~Singh, and C.~Guestrin, ``Near-optimal sensor placements in
  {G}aussian processes: Theory, efficient algorithms and empirical studies,''
  \emph{J. Mach. Learn. Res.}, vol.~9, pp. 235--284, 2008.

\bibitem[Cover and Thomas(2012)]{cover2012}
T.~M. Cover and J.~A. Thomas, \emph{Elements of Information Theory}.\hskip 1em
  plus 0.5em minus 0.4em\relax New York, NY: John Wiley \& Sons, 2012.

\bibitem[Foldes and Hammer(2005)]{foldes2005}
S.~Foldes and P.~L. Hammer, ``Submodularity, supermodularity, and higher-order
  monotonicities of pseudo-boolean functions,'' \emph{Mathematics of Operations
  Research}, vol.~30, no.~2, pp. 453--461, 2005.

\bibitem[Choi et~al.(2009)Choi, Brunet, and How]{choi2009tro}
H.-L. Choi, L.~Brunet, and J.~P. How, ``Consensus-based decentralized auctions
  for robust task allocation,'' \emph{{IEEE} Trans. Robotics}, vol.~25, no.~4,
  pp. 912--926, 2009.

\bibitem[Williams et~al.(2017)Williams, Gasparri, and Ulivi]{williams2017icra}
R.~K. Williams, A.~Gasparri, and G.~Ulivi, ``Decentralized matroid optimization
  for topology constraints in multi-robot allocation problems,'' in \emph{Proc.
  of the {IEEE} Intl. Conf. on Robot. and Autom.}, Singapore, May 2017.

\bibitem[Zhang and Vorobeychik(2016)]{zhang2016aaai}
H.~Zhang and Y.~Vorobeychik, ``Submodular optimization with routing
  constraints,'' in \emph{Assoc. for Adv. of Artif. Intell.}, 2016.

\bibitem[Charrow et~al.(2015)Charrow, Kahn, Patil, Liu, Goldberg, Abbeel,
  Michael, and Kumar]{charrow2015rss}
B.~Charrow, G.~Kahn, S.~Patil, S.~Liu, K.~Goldberg, P.~Abbeel, N.~Michael, and
  V.~Kumar, ``Information-theoretic planning with trajectory optimization for
  dense 3{D} mapping,'' in \emph{Proc. of Robot.: Sci. and Syst.}, Rome, Italy,
  Jul. 2015.

\bibitem[Indelman et~al.(2014)Indelman, Carlone, and
  Dellaert]{indelman2014icra}
V.~Indelman, L.~Carlone, and F.~Dellaert, ``Planning under uncertainty in the
  continuous domain: a generalized belief space approach,'' in \emph{Proc. of
  the {IEEE} Intl. Conf. on Robot. and Autom.}, Hong Kong, China, Jun. 2014.

\bibitem[Browne et~al.(2012)Browne, Powley, Whitehouse, Lucas, Cowling,
  Rohlfshagen, Tavener, Perez, Samothrakis, and Colton]{browne2012}
C.~Browne, E.~Powley, D.~Whitehouse, S.~Lucas, P.~I. Cowling, P.~Rohlfshagen,
  S.~Tavener, D.~Perez, S.~Samothrakis, and S.~Colton, ``A survey of {M}onte
  {C}arlo tree search methods,'' \emph{{IEEE} Trans. on Comput. Intell. and
  {AI} in Games}, vol.~4, no.~1, pp. 1--43, 2012.

\bibitem[Chaslot(2010)]{chaslot2010}
G.~Chaslot, ``Monte-{C}arlo tree search,'' Ph.D. dissertation, Universiteit
  Maastricht, 2010.

\bibitem[Ryan and Hedrick(2010)]{ryan2010ras}
A.~Ryan and J.~K. Hedrick, ``Particle filter based information-theoretic active
  sensing,'' \emph{Robot. Auton. Syst.}, vol.~58, no.~5, pp. 574--584, 2010.

\end{thebibliography}
}

\ifextended
\appendices

\clearpage
\section{The chain rule for derivatives
of set functions}
\label{appendix:chain_rule}

Set functions and their derivatives satisfy chain rules analogous to those
for entropy and mutual information~\citep[Theorem~2.5.1--2]{cover2012}.
We provide a general statement here which we will apply to first and
second derivatives of set functions.
\begin{lemma}[Chain rule for derivatives of set functions]
  \label{lemma:chain_rule}
  Consider sets $Y_1,\ldots,Y_n,X\subseteq\ground$, all disjoint.
  Then, writing the elements of $Y_n$ as $Y_n=\{y_{n,1},\ldots,y_{n,|Y_n|}\}$,
  the derivative of $\setfun$ can be rewritten in terms of derivatives with respect
  to the individual elements of $Y_n$ as
  \begin{align}
    \setfun(Y_1;\ldots;Y_n|X)
    = \sum_{i=1}^{|Y_n|} \setfun(Y_1;\ldots;Y_{n-1};y_{n,i}|Y_{n,1:i-1},X).
    \label{eq:chain_rule}
  \end{align}
\end{lemma}

\begin{proof}
  The proof follows by expanding the derivative \eqref{eq:derivative}, forming a
  telescoping sum, and rewriting the summands as individual derivatives:
  \begin{align}
    \setfun(Y_1;\ldots;Y_n|X)
    =&
    \setfun(Y_1;\ldots;Y_{n-1}|Y_n,X) \nonumber\\
     &\quad - \setfun(Y_1;\ldots;Y_{n-1}|X) \nonumber\\
    =&
    \sum_{i=1}^{|Y_n|}
    \big(
      \setfun(Y_1;\ldots;Y_{n-1}|Y_{n,1:i},X) \nonumber\\
     &\hspace{2em} - \setfun(Y_1;\ldots;Y_{n-1}|Y_{n,1:i-1},X)
    \big)
    \nonumber\\
    =&
    \sum_{i=1}^{|Y_n|}
    \setfun(Y_1;\ldots;Y_{n-1};y_{n,i}|Y_{n,1:i-1},X).
  \end{align}
\end{proof}

\section{Proof of Lemma~\ref{lemma:approximate_suboptimality}, suboptimality
of general assignments}
\label{appendix:approximate_suboptimality}

\begin{proof}
  This result follows typical methods for sequential submodular
  maximization
  with slight changes to assist in book-keeping:
  \begin{align}
    \setfun(X^\opt) &\leq \setfun(X^\mathrm{d}, X^\opt) \label{eq:line1}\\
               &=    \setfun(X^\mathrm{d}) + \sum_{i=1}^{\numrobots}
                     \setfun(x^\star_i | X^\star_{1:i-1}, X^\mathrm{d})
                     \label{eq:line2}\\
               &\leq \setfun(X^\mathrm{d}) + \sum_{i=1}^{\numrobots}
                     \setfun(x^\star_i | X^\mathrm{d}_{1:i-1})
                     \label{eq:line3}\\
               &\leq \setfun(X^\mathrm{d}) + \sum_{i=1}^{\numrobots}
                     \max_{x \in \block_i} \setfun(x|X^\mathrm{d}_{1:i-1})
                     \label{eq:linea}\\
               &\leq \setfun(X^\mathrm{d}) + \sum_{i=1}^{\numrobots}
                     \left(
                       \setfun(x^\mathrm{d}_i | X^\mathrm{d}_{1:i-1})
                       + \costgeneral_i(\setfun, x^\mathrm{d}_i, X^\mathrm{d}_{1:i-1})
                     \right)
                     \label{eq:line4}
                     \\
               &=    2\setfun(X^\mathrm{d})
                     + \sum_{i=1}^{\numrobots}
                     \costgeneral_i(\setfun, x^\mathrm{d}_i, X^\mathrm{d}_{1:i-1}).
                     \label{eq:line5}
  \end{align}
  Above,
  \eqref{eq:line1} follows from monotonicity;
  \eqref{eq:line2} expands a telescoping series;
  \eqref{eq:line3} follows from submodularity;
  \eqref{eq:linea} upper bounds the gains for the optimal decisions $x^\star_i$
  with the maximum marginal gains;
  \eqref{eq:line4} substitutes the expression for general suboptimality
  \eqref{eq:single_robot_suboptimality}
  thereby adding and subtracting the marginal gains for $x^\mathrm{d}_i$; and
  \eqref{eq:line5} collapses the telescoping series.
\end{proof}

\section{Proof of Theorem~\ref{theorem:main_bound}, suboptimality of
distributed planning}
\label{sec:proof_of_main_bound}

\begin{proof}
  Theorem~\ref{theorem:main_bound} consists of two parts,
  \eqref{eq:costs_bound} and \eqref{eq:weights_bound}.
  We prove each in turn.
  Since the costs in both equations involve sums over robots, both proofs
  analyze costs with respect to some robot $i\in\robots$.

  \subsubsection{Proof of Theorem~\ref{theorem:main_bound}, part 1
  \eqref{eq:costs_bound}}
  According to the standard greedy algorithm \eqref{eq:sequential_greedy}, robot
  $i$ would plan conditional on decisions by robots $\{1,\ldots,i-1\}$.
  However, in Alg.~\ref{alg:distributed} that robot instead plans conditional on
  decisions by
  a subset of these robots $\inneighbor_i\subseteq\{1,\ldots,i-1\}$ and ignores
  $\ignore_i=\{1,\ldots,i-1\}\setminus\inneighbor_i$.
  Recalling Lemma~\ref{lemma:approximate_suboptimality},
  we can write the suboptimality of decisions $X^\mathrm{d}$ in terms of a
  general cost $\costgeneral_i$.
  Let us now extract the cost of distributed planning $\distcost_i$ from
  this expression as follows:
  \begin{align}
    \begin{split}
      \costgeneral_i(\setfun, x^\mathrm{d}_i, X^\mathrm{d}_{1:i-1})
      &\leq
      \max_{x\in\block_i} \setfun(x|X^\mathrm{d}_{\inneighbor_i})
      - \setfun(x_i^\mathrm{d}|X^\mathrm{d}_{1:i-1})
      \\
      &=
      \costgeneral_i(\setfun, x^\mathrm{d}_i, X^\mathrm{d}_{\inneighbor_i})
      \\&\hspace{5ex}
      + \setfun(x_i^\mathrm{d}|X^\mathrm{d}_{\inneighbor_i})
      - \setfun(x_i^\mathrm{d}|X^\mathrm{d}_{1:i-1})
      \\
      &=
      \costgeneral_i(\setfun, x^\mathrm{d}_i, X^\mathrm{d}_{\inneighbor_i}) + \distcost_i.
    \end{split}
    \label{eq:substitute_distributed_cost}
  \end{align}
  Here, the first step follows by referring to the general cost
  model \eqref{eq:single_robot_suboptimality} and
  observing that
  $\max_{x\in\block_i} \setfun(x|X^\mathrm{d}_{1:i-1})
  \leq
  \max_{x\in\block_i} \setfun(x|X^\mathrm{d}_{\inneighbor_i})$
  due to submodularity.
  The second rewrites the cost in terms of decisions with respect to
  $X^\mathrm{d}_{\inneighbor_i}$; and the last substitutes
  the cost of distributed planning \eqref{eq:distributed_planning_cost}.

  To incorporate the cost of suboptimal planning $\plannercost_i$, observe that
  \begin{align}
    \begin{split}
      \costgeneral_i(\setfun, x^\mathrm{d}_i, X^\mathrm{d}_{\inneighbor_i})
      &=
      \costgeneral_i(\setfun, x^\mathrm{d}_i, X^\mathrm{d}_{\inneighbor_i})
      + \plannercost_i
      - \plannercost_i
      \\
      &=
      \plannercost_i
      + \costgeneral_i(\setfun, x^\mathrm{d}_i, X^\mathrm{d}_{\inneighbor_i})
      \\&\hspace{6ex}
      - \costgeneral_i(\approxsetfun_i, x^\mathrm{d}_i, X^\mathrm{d}_{\inneighbor_i})
    \end{split}
    \label{eq:substitute_planner_cost}
  \end{align}
  which follows from the definition of the planning cost
  \eqref{eq:planner_cost}.

  The cost of approximation of the objective $\objectivecost_i$
  upper bounds the difference of the last two terms in
  \eqref{eq:substitute_planner_cost}:
  \begin{align}
    \begin{split}
      \costgeneral_i (\setfun &, x^\mathrm{d}_i, X^\mathrm{d}_{\inneighbor_i})
      - \costgeneral_i(\approxsetfun_i, x^\mathrm{d}_i, X^\mathrm{d}_{\inneighbor_i})\\
      &=
      \approxsetfun_i(x_i^\mathrm{d} | X^\mathrm{d}_{\inneighbor_i})
      -
      \setfun(x_i^\mathrm{d} | X^\mathrm{d}_{\inneighbor_i})
      \\&
      \hspace{2em} +
      \max_{x \in \block_i}
      \setfun(x | X^\mathrm{d}_{\inneighbor_i})
      -
      \max_{x \in \block_i}
      \approxsetfun_i(x | X^\mathrm{d}_{\inneighbor_i})
      \\
      &\leq
      \approxsetfun_i(x_i^\mathrm{d} | X^\mathrm{d}_{\inneighbor_i})
      -
      \setfun(x_i^\mathrm{d} | X^\mathrm{d}_{\inneighbor_i})
      \\&
      \hspace{2em} +
      \setfun(\hat x | X^\mathrm{d}_{\inneighbor_i})
      -
      \approxsetfun_i(\hat x | X^\mathrm{d}_{\inneighbor_i}),
      \\&
      \hspace{4em}
      \text{for }
      \hat x \in \argmax \setfun(\hat x | X^\mathrm{d}_{\inneighbor_i})
      \\
      &\leq \objectivecost_i.
      \label{eq:substitute_objective_cost}
    \end{split}
  \end{align}
  The equality in \eqref{eq:substitute_objective_cost} follows by expanding and
  rearranging the costs \eqref{eq:single_robot_suboptimality} on the
  left-hand-side.
  The first inequality swaps the subtracted maximum for the approximate
  marginal gain $\approxsetfun_i$ at the point of the first maximum.
  Then, the second inequality uses the definition of the objective cost
  \eqref{eq:objective_cost} (maximum over- and under- approximation) to bound
  the two differences.

  Then, the expression for the costs in \eqref{eq:costs_bound}
  \begin{align}
    \costgeneral_i(\setfun, x^\mathrm{d}_i, X^\mathrm{d}_{1:i-1})
    &\leq
    \objectivecost_i
    + \plannercost_i
    + \distcost_i
    \label{eq:general_and_specific_costs}
  \end{align}
  follows by substituting the prior three equations into each other:
  \eqref{eq:substitute_objective_cost} into
  \eqref{eq:substitute_planner_cost} and the result into
  \eqref{eq:substitute_distributed_cost}.
  Finally, substituting this inequality \eqref{eq:general_and_specific_costs}
  into \eqref{eq:approximate_suboptimality} from
  Lemma~\ref{lemma:approximate_suboptimality}
  (on the suboptimality of general assignments)
  yields the desired bound \eqref{eq:costs_bound} which completes
  the first part of this proof.

  \subsubsection{Proof of Theorem~\ref{theorem:main_bound}, part 2
  \eqref{eq:weights_bound}}
  The second part of Theorem~\ref{theorem:main_bound} \eqref{eq:weights_bound}
  follows by referring to definition of $\distcost_i$ in
  \eqref{eq:distributed_planning_cost}, applying the chain rule
  \eqref{eq:chain_rule}, and substituting the definitions of the weights
  \eqref{eq:weights} and \eqref{eq:weight_by_component} in turn:
  \begin{align}
    \distcost_i
    &=
    - \setfun(x_i^\mathrm{d};X^\mathrm{d}_{\ignore_i}|X^\mathrm{d}_{\inneighbor_i})
    \\
    &=
    - \sum_{j \in \ignore_i}
    \setfun\big(
      x_i^\mathrm{d};x^\mathrm{d}_{j}|
      X^\mathrm{d}_{\inneighbor_i}, X^\mathrm{d}_{\ignore_i\cap\{1:j-1\}}
    \big)
    \\
    &\leq
    \sum_{j \in \ignore_i}
    \weights(i,j)
    \leq
    \sum_{j \in \ignore_i}
    \approxweights(i,j).
  \end{align}
  Then,~\eqref{eq:weights_bound} follows by summing over $\robots$.
  This completes this second and last part of the proof of
  Theorem~\ref{theorem:main_bound}.
\end{proof}

\section{Analysis for scaling to large numbers of robots}
\label{sec:scaling_analysis}

The analysis in this section establishes sufficient conditions for the cost of
distributed planning $\distcost$~\eqref{eq:distributed_planning_cost} for each
robot to be constant (in expectation) for planners with a fixed number of
sequential steps, independent of the number of robots.
Afterward, we discuss how this analysis relates to the design and analysis of
target tracking systems.

\subsection{Bounding expected inter-agent redundancy}
\label{sec:inter_agent_scaling}

Consider a distribution of robots and targets on $\real^n$ with at most $\alpha$
robots and $\beta$ targets on average per unit volume.
Then, assume that the channel capacities~\eqref{eq:target_capacities} between
each robot $i\in\robots$ and target $j\in\targets$ satisfy a non-increasing
upper bound
$\phi: \real_{\geq 0} \rightarrow \real_{\geq 0}$
(possibly in expectation) so that
$C_{i,j} \leq \phi(||\robotposition_i - \targetposition_j||_2)$
where $\robotposition_i$ and $\targetposition_j$ are the robot position and
target \emph{mean} position in $\real^n$.

Now, consider the expectation of the total weight associated with robot
$i\in\robots$
considering only robots and targets on an
$n$-ball with radius
$R/2$ centered around $\robotposition_i$
which we write as
$\E_{R/2}[\sum_{j\in \robots\setminus\{i\}} \approxweights(i,j)]$.
Consider also the expectation for targets distributed within a radius $R$
around $i$ and robots on balls with the same radius centered on each target.
The intersection of these balls is a ball around $i$ with radius $R/2$
so that the latter expectation produces an upper bound on the former.
Given the expression for $\approxweights$ in terms of channel
capacities~\eqref{eq:weight_by_component}, and designating the zero-centered
ball with radius $R$ as $B_R$ we can write this inequality as:
\begin{align}
  &\E_{B_{R/2}}\bigg[\sum_{j\in \robots\setminus\{i\}} \approxweights(i,j)\bigg]
  \nonumber \\
  &\leq
  \int_{B_R}\int_{B_R}
  \alpha\beta
  \min(\phi(||\vec{x}||_2), \phi(||\vec{y}||_2))
  \dd{\vec{x}}\dd{\vec{y}}.
  \intertext{By integrating over the surface of each ball
  (each an $(n-1)$-sphere with surface area $S_{n-1}$)}
  &=
  \alpha\beta
  \!
  \int_0^{R}
  \!
  \int_0^{R}
  {S_{n-1}}^2
  {r_1}^{n-1}{r_2}^{n-1}
  \min(\phi(r_1), \phi(r_2))
  \dd{r_1}\dd{r_2}.
  \intertext{Given that $\phi$ is non-increasing, separating the minimum
  produces:}
  &=
  \alpha\beta
  {S_{n-1}}^2
  \left(
    \int_0^{R}
    \int_0^{r_2}
    {r_1}^{n-1}{r_2}^{n-1}
    \phi(r_2)
    \dd{r_1} \dd{r_2}
  \right.
    \nonumber\\
  &\hspace{15ex}
  \left.
    +
    \int_0^{R}
    \int_{r_2}^{R}
    {r_1}^{n-1}{r_2}^{n-1}
    \phi(r_1)
    \dd{r_1} \dd{r_2}
  \right),
  \\
  \intertext{and by swapping the bounds of the second integral, combining, and
  evaluating the inner integral, we get:}
  &=
  \alpha\beta
  {S_{n-1}}^2
  \left(
    \int_0^{R}
    \int_0^{r_2}
    {r_1}^{n-1}{r_2}^{n-1}
    \phi(r_2) \dd{r_1} \dd{r_2}
  \right.
    \nonumber\\
  &\hspace{15ex}
  \left.
    +
    \int_0^{R}
    \int_0^{r_1}
    {r_1}^{n-1}{r_2}^{n-1}
    \phi(r_1) \dd{r_2} \dd{r_1}
  \right)
  \\
  &=
  2
  \alpha\beta
  {S_{n-1}}^2
  \int_0^{R}
  \int_0^{r_1}
  {r_1}^{n-1}{r_2}^{n-1}
  \phi(r_1) \dd{r_2} \dd{r_1}
  \\
  &=
  \frac{%
    2
    \alpha\beta
    {S_{n-1}}^2
  }{n}
  \int_0^{R}
  {r_1}^{2n-1} \phi(r_1)
  \dd{r_1}.
  \label{eq:target_tracking_scaling}
\end{align}
The above integral~\eqref{eq:target_tracking_scaling} converges in the limit
if $\phi \in O(1/x^{2n+\epsilon})$.
Most relevantly, for a plane, this
condition comes to $\phi \in O(1/x^{4+\epsilon})$.\footnote{%
  This requirement on interactions between
  \emph{robots and targets} is stricter than the equivalent one
  \emph{between robots}~\citep[Sec.~VI.C]{corah2018cdc} as
  interactions between robots must decrease as $O(1/x^{n+\epsilon})$.
}

\subsection{Scaling and sensor models}

The sensitivity to how quickly interactions between robots and targets fall off
motivates attention to sensing design and modeling to prevent distributed
planning from performing poorly for large numbers of robots or else requiring
additional computation time.
For example, considering additive Gaussian noise with a standard deviation
proportional to distance
(as is common in range sensing models~\citep{charrow2014ijrr})
mutual information does not fall off quickly enough as is evident from the
capacity of Gaussian channels~\citep[Chap.~9]{cover2012}.

At the same time, robots in realizable systems cannot obtain and process
observations of unbounded numbers of targets, and features such as a maximum
sensor range (as we use in the results) can model such limits.
Still, the observation of whether a target is within range provides some
information.
Introducing a narrow tails assumption on the target filters---such as that the
\emph{tails approach zero exponentially}---in combination with a maximum sensing
range ensures that $\phi$ decreases sufficiently quickly.
However, due to this sensitivity to the tails,
the scaling behavior~\eqref{eq:target_tracking_scaling} may be difficult to
estimate, even when known to be bounded.

\fi

\end{document}